\documentclass{article}

\usepackage{mathrsfs}
\usepackage{amssymb}
\usepackage{amsmath}
\usepackage{amsthm}

\usepackage{amsfonts}
\usepackage{microtype}
\usepackage{graphicx}
\usepackage{subfigure}

\newtheorem{theorem}{Theorem}
\newtheorem{lemma}{Lemma}
\newtheorem{corollary}{Corollary}
\newtheorem{remark}{Remark}
\newtheorem{definition}{Definition} 
\usepackage{booktabs} 

\usepackage{hyperref}
\usepackage{xcolor}
\usepackage{enumitem}

\usepackage[accepted]{icml2021}

\newcommand{\Name}{SPADE} 

\icmltitlerunning{{\Name}: A Spectral Method for Black-Box Adversarial Robustness Evaluation}

\begin{document}

\twocolumn[
\icmltitle{{\Name}: A \underline{Sp}ectral Method for Black-Box \underline{Ad}versarial Robustness \underline{E}valuation}



\icmlsetsymbol{equal}{*}

\begin{icmlauthorlist}
\icmlauthor{Wuxinlin Cheng}{equal,to}
\icmlauthor{Chenhui Deng}{equal,goo}
\icmlauthor{Zhiqiang Zhao}{equal,to}
\icmlauthor{Yaohui Cai}{goo}
\icmlauthor{Zhiru Zhang}{goo}
\icmlauthor{Zhuo Feng}{to}
\end{icmlauthorlist}

\icmlaffiliation{to}{Stevens Institute of Technology, New Jersey, USA}
\icmlaffiliation{goo}{Cornell University, New York, USA}

\icmlcorrespondingauthor{Zhuo Feng}{zhuo.feng@stevens.edu}
\icmlcorrespondingauthor{Zhiru Zhang}{zhiruz@cornell.edu}

\icmlkeywords{Machine Learning, ICML}

\vskip 0.3in
]



\printAffiliationsAndNotice{\icmlEqualContribution} 

\begin{abstract}
 A black-box spectral method is introduced for evaluating the adversarial robustness of a given machine learning (ML) model. Our approach, named SPADE, exploits  bijective distance mapping between the input/output graphs constructed for approximating the manifolds corresponding to the input/output data. By leveraging the generalized Courant-Fischer theorem, we propose a SPADE score for evaluating the adversarial robustness of a given model, which is proved to be an upper bound of the best Lipschitz constant under the manifold setting. To  reveal the most non-robust data samples highly vulnerable to adversarial attacks, we develop a spectral graph embedding procedure leveraging dominant generalized eigenvectors. This embedding step allows assigning each data sample a robustness score that can be further harnessed for more effective adversarial training. 
 Our experiments show the proposed SPADE method leads to promising empirical results for neural network models that are adversarially trained with the MNIST and CIFAR-10 data sets. 
\end{abstract}

\section{Introduction}
\label{sec:intro}
 Recent research efforts have demonstrated the evident lack of robustness in state-of-the-art machine learning (ML) models---e.g., a visually imperceptible adversarial image can be crafted via an optimization procedure to mislead a well-trained deep neural network (DNN)  \cite{szegedy2013intriguing, goodfellow2014explaining}. Consequently, it is becoming increasingly important to effectively assess and improve the adversarial robustness of ML models for safety-critical applications, such as autonomous driving systems. To this end, a variety of white-box approaches has been proposed. For instance, study in \cite{szegedy2013intriguing} proposed a layer-wise global Lipschitz constant estimation approach, which provides a loose bound on robustness evaluation;    \cite{hein2017formal} introduced a method for assessing the lower bound of model robustness based on  local Lipschitz continuous condition for a multilayer perceptron (MLP) with a
single hidden layer; \cite{weng2018evaluating} proposed a method for estimating local Lipschitz constant based on extreme value theory. However, most existing adversarial robustness evaluation frameworks are based on white-box methods which assume the model parameters are given in advance. For example, the recent CLEVER algorithm for adversarial robustness evaluation \cite{weng2018evaluating} requires full access to the gradient information of a given neural network for estimating a universal lower
bound on the minimal distortion required to craft an adversarial example from an original one.

This work introduces SPADE \footnote{The SPADE source code is available at  \href{https://github.com/Feng-Research/SPADE}{github.com/Feng-Research/SPADE}.}, a black-box method for evaluating 
adversarial robustness by only using the input (features) and output vectors of the ML model. Essentially, our method evaluates adversarial robustness through checking if there exist two nearby input data samples that can be mapped to very distant output ones by the underlying function of the ML model; if so, we have a large distance mapping distortion (DMD), which implies potentially poor adversarial robustness since a small perturbation applied to these inputs can lead to rather significant changes on the output side. To allow meaningful distance comparisons of  input/output data samples in a high-dimensional space, our approach leverages graph-based manifolds and focuses on resistance distance metric for adversarial robustness evaluations. The main contributions of this work are as follows:

\noindent $\bullet$ To our knowledge, we are the first to introduce a black-box spectral method (SPADE) for adversarial robustness evaluation of an ML model by examining the bijective distance mappings between the input/output graph-based manifolds. 

\noindent $\bullet$ We show that the largest generalized eigenvalue (i.e., SPADE score) computed with the input/output graph Laplacians can be a good surrogate (upper bound) for the best Lipschitz constant of the  underlying function, which thus can be leveraged for quantifying the adversarial robustness.

\noindent $\bullet$  We propose   a spectral graph embedding scheme  leveraging the generalized Courant-Fischer theorem    for estimating the  robustness of each data point:     a data point  with a larger SPADE score  means it may contain a greater amount of non-robust  features, and thus can be more vulnerable to adversarial perturbations.

\noindent $\bullet$  We show that the SPADE score of an ML model can be directly used as a black-box metric for   quantifying its adversarial robustness. Moreover,  by taking advantage of   the SPADE scores of input data samples,  {existing methods for  adversarial training  can be {further} improved,  achieving state-of-the-art performance.} 

\section{Background}
\label{sec:background}
\subsection{Spectral Graph Theory}

For an undirected graph $G=(V,E,w)$, $V$ denotes a set of nodes (vertices), $E$ denotes a set of (undirected) edges, and $w$ denotes the associated edge weights. The graph adjacency matrix can be defined as:
\begin{equation}\label{di_adjacency}
{A}(i,j)=\begin{cases}
w(i,j) & \text{ if } (i,j)\in E \\
0 & \text{otherwise }
\end{cases}
\vspace{-5pt}
\end{equation}

Let $D$ denote the diagonal matrix with ${D}(i,i)$ being equal to the (weighted) degree of node $i$. The graph Laplacian matrix can be constructed by ${L={D}-{A}}$.

\iftrue
\begin{lemma}\label{lamma:eig}
(Courant-Fischer Minimax Theorem) The $k$-th largest eigenvalue of the Laplacian matrix $L \in \mathbb{R}^{|V|\times|V|}$ can be computed as follows:
\begin{equation}
    \lambda_k(L) = \min_{dim(U)=k}\,{\max_{\substack{x \in U \\ x \neq 0}}{\frac{x^\top Lx}{x^\top x}}}
\end{equation}
\end{lemma}
Lemma \ref{lamma:eig} describes the Courant-Fischer Minimax Theorem \cite{golub2013matrix} for computing the spectrum of the Laplacian matrix $L$. 

A more general form for Lemma \ref{lamma:eig} is referred as the generalized Courant-Fischer Minimax Theorem \cite{golub2013matrix}, which can be described as follows:
\fi
\begin{lemma}\label{lamma:generalized_eig}
(The Generalized Courant-Fischer Minimax Theorem) Given two Laplacian matrices $L_X \in \mathbb{R}^{|V|\times|V|}$ and $L_Y \in \mathbb{R}^{|V|\times|V|}$ such that $null(L_Y) \subseteq null(L_X)$, the $k$-th largest eigenvalue of $L_Y^{+}L_X$ can be computed as follows under the condition of $1 \leq k \leq rank(L_Y)$:
\begin{equation}
    \lambda_k(L_Y^{+}L_X) = \min_{\substack{dim(U)=k \\ U \bot null(L_Y)}}\,{\max_{x \in U }{\frac{x^\top L_Xx}{x^\top L_Yx}}}
\end{equation}
\end{lemma}

\subsection{Adversarially Robust Machine Learning}
Machine learning has been increasingly deployed in the safety- and security-sensitive applications, such as vision for autonomous cars, malware detection, and face recognition \cite{biggio2013security,kloft2010online}. There is an active body of research on adversarial ML, which attempts to understand and improve the robustness of the ML models. 
For example, adversarial attack aims to mislead the ML techniques by supplying the deceptive inputs such as input samples with perturbations \cite{goodfellow2018making, fawzi2018analysis}, which are commonly known as adversarial examples. It has been shown that state-of-the-art ML techniques are highly vulnerable to adversarial input samples during both training and inference \cite{ szegedy2013intriguing, nguyen2015deep, moosavi2016deepfool}. Hence resistance to adversarially chosen inputs is becoming a very important goal for designing ML models \cite{madry2018towards, barreno2010security}. 

There are also a number of defending methods proposed to mitigate the effects of adversarial attacks, which can be broadly divided into reactive and proactive defence. Reactive defence focuses on the detection of the adversarial examples from the model inputs \cite{feinman2017detecting, metzen2017detecting, xu2018feature, yang2020ml}. Proactive defence, on the other hand, tries to improve the  robustness of the models so they are not easily fooled by the adversarial examples \cite{gu2014towards, cisse2017parseval, shaham2018understanding,liu2018adv, xu2019topology, feng2019graph, jin2020graph}. 
These techniques usually make use of model parameter regularisation and robust optimization. While different defense mechanisms may be effective against certain classes of attacks, none of them
are deemed as a one-stop solution to achieving adversarial robustness. 

\subsection{Methods for  Adversarial Robustness Evaluation}
Although there are flourishing attack and defense approaches through adversarial examples, little progress has been made towards an attack-agnostic, black-box, and computationally affordable quantification of robustness level. For example, most existing approaches measure the robustness of a neural network via the attack success rate or the distortion of the adversarial examples yielded from certain attacks, such as the fast gradient sign method (FGSM)~\cite{goodfellow2014explaining, Kurakin2016}, Carlini \& Wagner’s attack (CW)~\cite{CW}, and projected gradient descent (PGD)~\cite{madry2018towards}. As \cite{weng2018evaluating} elaborated, for a given dataset and the corresponding adversarial examples yielded from an attack algorithm, the success rate of attack and the distortion of adversarial examples are treated as robustness metrics. Due to the entanglement between network robustness and the attack algorithm, such kinds of robustness measurements can cause biased analysis. Moreover, attack capabilities also limit the analysis. In contrast, our proposed robustness metric is attack-agnostic and thus avoids the above issues.

Recently, \cite{weng2018evaluating} proposed a robustness metric called CLEVER score
that consists of two major steps to compute. The first step is computing the cross Lipschitz constant $L^{j}_{q,x_{0}}$, which is defined as the maximum $\left\|\nabla g\left(x\right)\right\|_{q}$, where $p$ is the perturbation norm, $q=\dfrac{p}{p-1}$, $g\left(x\right)=f_{c}\left(x\right)-f_{j}\left(x\right)$, and the $f$ is a neural network classifier. 
Second, the location estimate, which is the maximum likelihood estimation of location parameter of reverse Weibull distribution on maximum $\left\|\nabla g\left(x^{\left(i,k\right)}\right)\right\|_{q}$ in each batch, is used as an estimation for the local cross Lipschitz constant (i.e., the CLEVER score).
The robustness metric CLEVER score is a reasonably effective estimator of the lower bound of minimum distortion. 
It can roughly indicate the best possible attack in terms of distortion. 
However, it is important to note that CLEVER falls into the white-box measurement category. It requires backpropagation where weights between different layers and activation functions at each layer are needed to calculate $\nabla g\left(x\right)$, which is computationally costly. Different from the CLEVER score, our metric targets the black-box measurement of robustness and has a lower computational cost.

\vspace{-7pt}
\section{The SPADE Robustness Metric} 
\label{sec:spade}
\begin{figure*}[ht]
\vskip 0.2in
\begin{center}
\centerline{\includegraphics[width=2\columnwidth]{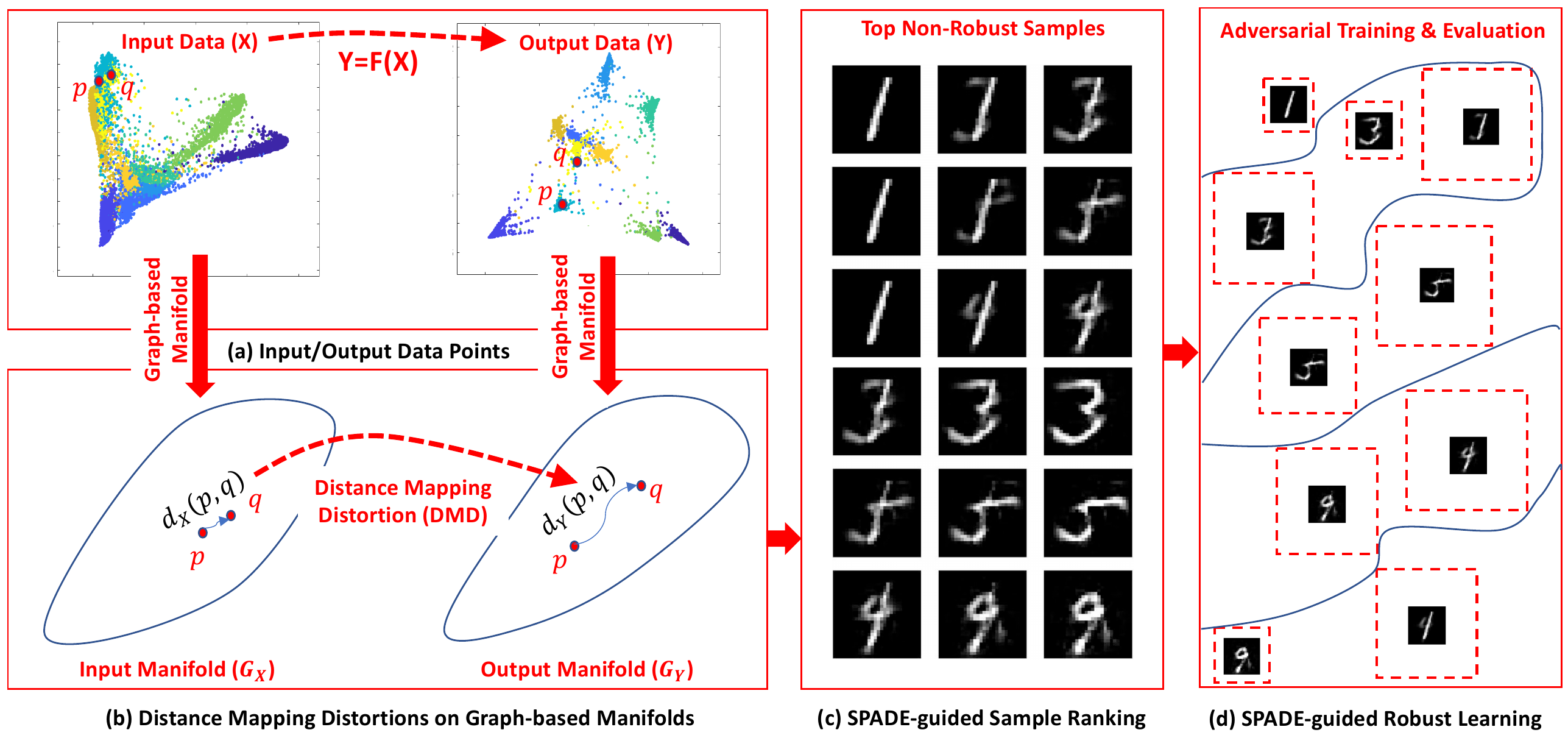}}
\caption{Overview of the proposed method. (a) Given  bijective input ($X$) and output ($Y$) data samples, SPADE first constructs graph-based manifolds. (b) SPADE  exploits   distance mapping distortions (DMDs) on  manifolds for adversarial robustness evaluation. (c) Each data sample is given a SPADE score to reflect its level of non-robustness. (d) Applications for SPADE-guided adversarially-robust  ML.}
\vspace{-8pt}
\label{fig:overview}
\end{center}
\vspace{-15pt}
\end{figure*}
Figure \ref{fig:overview} shows an overview of SPADE, our spectral method for black-box adversarial robustness evaluation. There are four key steps in our proposed approach: \textbf{(a)}  We first construct graph-based manifolds for both input and output data of a given ML model. \textbf{(b)} We then compute the {SPADE score}  for measuring the robustness of the ML model based on bijective distance mapping under the manifold setting. \textbf{(c)} We further extend the {SPADE score} to quantify the robustness of each input data sample. \textbf{(d)} We also develop SPADE-guided methods for adversarial training and robustness evaluation. As discussed in Section \ref{sec:app}, the SPADE-guided adversarial training can be done by adaptively setting the size  of the norm-bounded perturbation  for each data sample according to its   SPADE score, such that stronger defenses can be set up   for  more vulnerable data samples.  

\subsection{Graph-based Manifold Construction}
\label{sec:mnfold_construt}
In this work, we assume that the input/output data lie near a low-dimensional manifold~\cite{fefferman2016testing}. We analyze the adversarial robustness of an ML model by transforming its input/output data into a graph, which is a discrete approximation to the underlying manifold. More concretely, consider a given model (e.g., neural networks) that maps a reshaped $M$-dimensional input feature (e.g., image) $x_i\in {\mathbb{R} ^{M}}$ to a $D$-dimensional output vector $y_i\in {\mathbb{R} ^{D}}$ through a black-box 
mapping function $y_i=F(x_i)$. SPADE leverages the k-nearest-neighbor (kNN) algorithm to construct the input (output) graph for input (output) data points, as illustrated by $G_X$ ($G_Y$) in Figure \ref{fig:overview}. 

Similar to a recent work that exploits graph-based manifolds for the topology analysis of DNNs \cite{naitzat2020topology}, we only consider unweighted graphs  in this work (namely, each edge has a unit weight). In addition, we choose a proper $k$ value such that the input/output graph is connected. It is worth noting that a na\"ive implementation of kNN requires $O(|V|^2)$ time complexity to construct the graph with $O(|V|)$ nodes, which cannot scale to the datasets with millions of data points. Instead, we can leverage 
an extension of the probabilistic skip list structure 
to approximate kNN graphs with the complexity of $O(|V|\log|V|)$~\cite{malkov2018efficient}.

\subsection{The SPADE Score for ML Models}
\label{sec:model_metric}

After constructing graph-based manifolds for inputs and outputs, we can analyze the adversarial robustness under the manifold setting  through calculating the following metric. 

\begin{definition}
\label{gamma}
The \textbf{distance  mapping distortion (DMD)}  $\gamma^F(p,q)$ for a  node pair $(p, q)$ through a function $Y=F(X)$ is defined below, where $d_X(p,q)$ and $d_Y(p,q)$ denote the distances between nodes $p$ and $q$ on the input and output graphs, respectively.
\begin{equation}\label{formula_dmd}
\gamma^F(p,q)\overset{\mathrm{def}}{=}\frac{d_Y(p,q)}{d_X(p,q)}
\end{equation}
\end{definition}


\begin{remark}
 When $d_X(p,q)\to 0 $ (i.e., small input perturbation), the  DMD metric $\gamma^F(p,q)$   can be  regarded as a surrogate  for the  gradient of the function $Y=F(X)$ under the manifold setting.
\end{remark}

 Intuitively,   the maximum DMD ($\gamma^F_{max}$) value obtained via exhaustively  searching over    all node pairs  can be    exploited for   estimating the  maximum distance change  on the output  graph (manifold) due to a small distance perturbation on the input graph (manifold), which therefore allows  evaluating the adversarial robustness of a given function (model). 
 
 Unlike existing adversarial robustness evaluation methods (e.g., \cite{szegedy2013intriguing, goodfellow2014explaining,hein2017formal, weng2018evaluating}), {the proposed DMD metric for adversarial robustness evaluation }  does not just target specific types of adversarial attacks or require full access to the underlying model parameters. 
 Instead, our metric can be {conveniently obtained} by only exploiting input/output data manifolds. 
 In addition, identifying and subsequently correcting the most problematic data samples, the ones that have relatively large DMD values and thus  will potentially lead to poor adversarial robustness, will allow training much more adversarially robust models. 

\subsubsection{Computing $\gamma^F_{max}$  via  Resistance Distance }
\textbf{Geodesic distance.} Pairwise distance calculations on the manifold will be  key to estimating  $\gamma^F_{max}$. To this end, the geodesic distance metric is arguably the most natural choice: for graph-based manifold problems  the shortest-path distance metric have been   adopted for approximating the geodesic distance on a  manifold in  nonlinear dimensionality reduction and neural network topological  analysis  \cite{tenenbaum2000global,naitzat2020topology}. However, to exhaustively search for   $\gamma^F_{max}$ will require computing all-pairs shortest-paths   between $N$ input (output) data points, which can be prohibitively expensive even when taking advantage of the state-of-the-art randomized method \cite{williams2018faster}.

\textbf{Resistance distance.} To avoid staggering computational cost, we propose to compute $\gamma^F_{max}$ using  effective-resistance distances. Although both geodesic distances and effective-resistances distances are legitimate notions of distances between the nodes on a graph, the latter has been extensively studied in modern spectral graph theory and  found  close connections to many important problems, such as the cover and commute time of random walks \cite{chandra1996electrical}, the number of spanning trees of a graph, etc. 
 \begin{lemma}\label{lemma_resistance}
 The effective-resistance distance $d^\textit{eff}({p,q})$ between any two nodes $p$ and $q$ for an   $N$-node undirected and  connected graph $G=(V,E)$ satisfies:
\begin{equation}\label{formula_Reff}
d^\textit{eff}({p,q})=e_{p,q}^\top L_G^+ e_{p,q}=\|U_N^\top e_{p,q}\|^2_2
\end{equation}
 where ${{e_{p,q}=e_{p}-e_{q}}}$, ${{e_{p}\in {\mathbb{R} ^{N}}}}$ denotes the standard basis vector with  the $p$-th element being $1$ and others being $0$, $L_G^+\in {\mathbb{R} ^{N\times N}}$ denotes the Moore–Penrose pseudoinverse of the graph Laplacian matrix $L_G\in {\mathbb{R} ^{N\times N}}$, and $U_N$ denotes the eigensubspace   matrix including $N-1$ nontrivial weighted Laplacian eigenvectors:
 \begin{equation}\label{formula_eigen}
U_N=\left[\frac{u_2}{\sqrt {\sigma_2}},..., \frac{u_N}{\sqrt {\sigma_N}}\right]\in {\mathbb{R} ^{N\times (N-1)}}
\end{equation}
where $0=\sigma_1<\sigma_2,...,\le\sigma_N$  denote the ascending  eigenvalues corresponding to their eigenvectors $u_1,...,u_N$.
  \end{lemma}
 \begin{lemma}\label{lemma_geodesic}
 The effective-resistance distance $d^\textit{eff}({p,q})$ and geodesic distance $d^\textit{geo}({p,q})$  between any two nodes $p$ and $q$ for an   $N$-node undirected  connected graph satisfies:
\begin{enumerate}
\item $d^\textit{eff}({p,q})=d^\textit{geo}({p,q})$ if there is only one path between   nodes $p$ and $q$;
\item $d^\textit{eff}({p,q})<d^\textit{geo}({p,q})$ otherwise.
\end{enumerate}
\end{lemma}
Lemma \ref{lemma_geodesic}   implies that   $d^\textit{eff}({p,q})=d^\textit{geo}({p,q})$ will always be valid  for  trees,  since there will be only   one path between any pair of two nodes in a tree. For general graphs, the resistance distance $d^\textit{eff}({p,q})$ is  bounded by the geodesic distance $d^\textit{geo}({p,q})$.

By leveraging resistance distance, we can avoid enumerating all node pairs for calculating $\gamma^F_{max}$ by solving the following combinatorial optimization problem:
\begin{equation}\label{DMD_opt}
\max {\gamma^F}=\mathop{\max_{\forall p, q\in V}}_{p\neq q}\frac{e_{p,q}^\top L_Y^+ e_{p,q}}{e_{p,q}^\top L_X^+ e_{p,q}}
\end{equation}

However, since $e_{p,q}$ is a discrete vector, the above combinatorial optimization problem has a super-linear complexity: approximately finding $\gamma^F_{max}$ via computing   all-pair effective-resistance distances  can be  achieved by leveraging Johnson–Lindenstrauss lemma \cite{spielman2011graph}. To avoid the  high computational complexity of solving (\ref{DMD_opt}),  the following SPADE score is proposed  for  estimating the upper bound of $\gamma^F_{max}$, which can be computed in   nearly-linear time leveraging recent fast Laplacian solvers \cite{koutis2010approaching,kyng2016approximate}.

\subsubsection{Estimating $\gamma^F_{max}$ via SPADE Score}

\begin{definition}
Denoting $L_X$ ($L_Y$)   the Laplacian matrix of the input (output) graph $G_X$ ($G_Y$),  the \textbf{SPADE score} of a function (model) $Y=F(X)$ is defined as:
\begin{equation}\label{formula_spade}
\textbf{SPADE}^F\overset{\mathrm{def}}{=}\lambda_{max}( L_Y^+L_X )
\end{equation}
\end{definition}

\begin{theorem}
 \label{theory_eigenvalue}
 When  computing $\gamma^F_{max}$ via effective-resistance distance, the \textbf{SPADE score} is an upper bound of $\gamma^F_{max}$.
\end{theorem}
The proof for Theorem \ref{theory_eigenvalue} is available in the Appendix.

\begin{definition} 
Given two metric spaces $(X, dist_X)$ and $(Y, dist_Y)$, where $dist_X$ and $dist_Y$ denote the distance metrics on the sets $X$ and $Y$, respectively, a function $Y=F(X)$ is called {Lipschitz continuous} if there exists a real constant $K\ge0$ such that for all $x_i$, $x_j\in X$:
\begin{equation}
    dist_Y(F(x_i),F(x_j))\le K dist_X(x_i,x_j),
\end{equation}
where $K$ is called the Lipschitz constant for the function $F$. The smallest Lipschitz constant denoted by $K^*$ is called the \textbf{best Lipschitz constant}.
\end{definition}

\begin{corollary}
\label{l_bound}
 Let the resistance distance be the distance metric, we have:
 \begin{equation}
 \lambda_{max}( L_Y^+L_X )\ge K^*\ge\gamma^F_{max}
 \end{equation}
\end{corollary}
 
Corollary \ref{l_bound} indicates that the SPADE score is also an upper bound of the best Lipschitz constant $K^*$ under the manifold setting. A greater SPADE score of a function (model) implies a  worse adversarial robustness, since the output  will be more sensitive to small input perturbations. Thus, we can use the SPADE score to quantify the robustness of a given ML model. We empirically show in Section \ref{spade_metric_eval} that a more robust model has a smaller SPADE score compared against non-robust models, which confirms the efficacy of our proposed approach.

\subsection{The SPADE Score for Input Data Samples}
\label{sec:data_metric}

Apart from proposing a metric for evaluating the robustness of machine learning models, we further develop a metric score for revealing the robustness level of each input {data sample}.   Consequently, we can utilize the {sample robustness}   score for  ML  applications discussed in Section \ref{sec:app}.

To measure the robustness per input data sample (i.e., per node), we first measure the robustness of node pairs following the notion of DMD defined in Section~\ref{sec:model_metric}.

\begin{definition}
\label{node_pair}
 A node pair $(p,q)$ is non-robust  if it has a large distance mapping distortion (e.g., $\gamma^F(p,q)\approx\gamma^F_{max}$).
\end{definition}



Intuitively, a non-robust node pair consists of nodes that are adjacent in the input graph $G_X$ but far apart in the output graph $G_Y$. To effectively reveal the non-robust node pairs, we introduce the cut mapping distortion metric as follows: 

\begin{definition}
For two graphs $G_X$ and $G_Y$ that share the same node set $V$, let $S \subset V$ denote a node subset and $\bar{S}$ denote the complement of $S$. Also let $cut_G(S, \bar{S})$ denote the number of edges crossing $S$ and $\bar{S}$ in graph $G$. The \textbf{cut mapping distortion (CMD)} $\zeta (S)$ of node subset $S$ is defined as:
\begin{equation}\label{cmd}
\zeta (S)\overset{\mathrm{def}}{=}\frac{cut_{G_Y}(S,  \bar{S}  )}{cut_{G_X}(S,  \bar{S}  )}
\end{equation}
\end{definition}

\begin{figure}[ht]
\vskip 0.2in
\begin{center}
\centerline{\includegraphics[width=0.99\columnwidth]{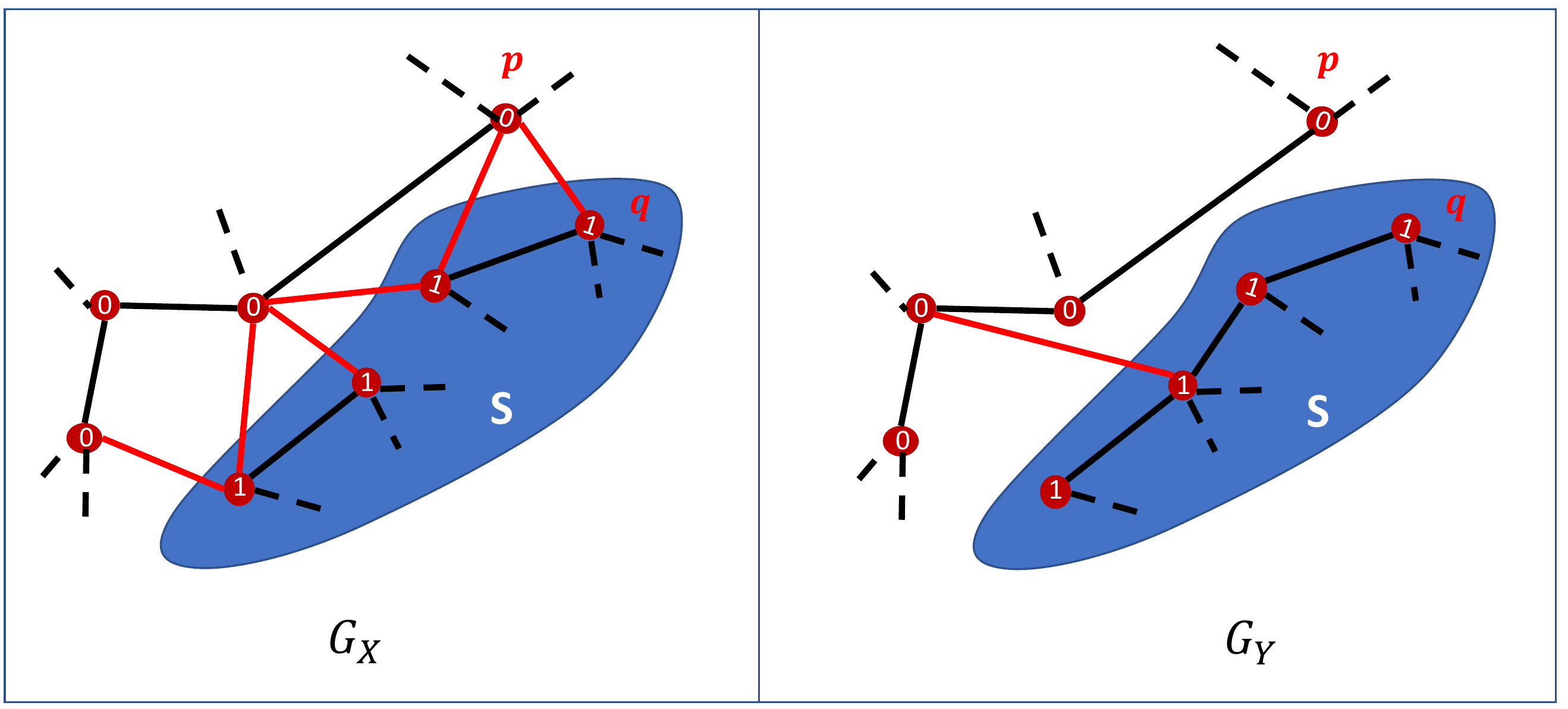}}
\caption{A node coloring vector assigns each node  an integer $0$ or $1$ to define the node subset $S$. The cut mapping distortion of $S$ can be computed by: $\zeta (S)=\frac{1}{6}$.}
\label{fig:cmd}
\end{center}
\vskip -0.2in
\end{figure}

A small CMD score indicates that the node pairs crossing the boundary of $S$ are likely to have small distances in $G_X$ but rather large distances in $G_Y$. As shown in Figure \ref{fig:cmd}, the node subset $S$ has six edges crossing the boundary in $G_X$ but only one in $G_Y$; as a result, with a high probability the node pairs crossing the boundary  will have much smaller effective-resistance or geodesic (shortest-path) distances in $G_X$ than $G_Y$. For example, nodes $p$ and $q$ are adjacent in $G_X$, while they have a large distance in $G_Y$ (the shortest-path distance is five).

\begin{theorem}\label{theory_cut}
Let $L_X$ and $L_Y$ denote the Laplacian matrices of input and output graphs, respectively. The following inequality holds for the minimum CMD $\zeta_{min}$:
\vspace{-4pt}
\begin{equation}\label{max_cut}
\zeta_{min}=\mathop{\min_{\forall S\subset V}} \zeta (S) \ge\frac{1}{\lambda_{max}( L_Y^+L_X )}
\vspace{-4pt}
\end{equation} 
\end{theorem}

The proof is available in the Appendix. Theorem \ref{theory_cut} shows the connection between the maximum generalized eigenvalue $\lambda_{max}( L_Y^+L_X )$ and $\zeta_{min}$, motivating us to exploit the largest generalized eigenvalues and their corresponding eigenvectors to measure the robustness of node pairs.

\paragraph{Embedding $G_X$ with generalized eigenpairs.} Specifically, we first compute the weighted eigensubspace matrix $V_r\in {\mathbb{R} ^{N\times r}}$ for spectral embedding on $G_X$ with $N$ nodes:
\vspace{-4pt}
\begin{equation}\label{subspace2}
V_r\overset{\mathrm{def}}{=}\left[ {v_1}{\sqrt {\lambda_1}},...,  {v_r}{\sqrt {\lambda_r }}\right],
\vspace{-4pt}
\end{equation}
where $\lambda_1, \lambda_2, ..., \lambda_r$ represent the first $r$ largest eigenvalues of $L_Y^+L_X$ and $v_1, v_2, ..., v_r$ are the corresponding eigenvectors. To this end, the input graph $G_X$ can be embedded using $V_r$ such that each node is associated with an $r$-dimensional embedding vector. Subsequently, we can  quantify the robustness of an edge $(p ,q) \in E_X$ via measuring the spectral embedding distance of its two end nodes  $p$ and $q$. Formally, we have the following definition:
 \begin{definition}
The  \textbf{edge SPADE score} is defined for  any edge $(p,q)\in E_X$   as follows:
\begin{equation}\label{edge_len}
    \textbf{SPADE}^F(p,q)\overset{\mathrm{def}}{=}\|V_r^\top e_{p,q}\|_2^2.
\vspace{-10pt}
\end{equation}
\vspace{-10pt}
\end{definition}
\begin{theorem}
 \label{theory_embed}
Denote the first $r$ dominant generalized eigenvectors of $L_X L_Y^+$ by $u_1, u_2, ..., u_r$. If an edge $(p,q)$  is  dominantly aligned with one  dominant eigenvector   $u_k$, where $1 \le k \le r$, the following holds:
\begin{equation}\label{edge_embed0}
(u^\top_i e_{p,q})^2 \approx \begin{cases}
\alpha_k^2 \gg 0 & \text{ if } (i=k) \\
0 & \text{ if } (i\neq k).
\end{cases}
\vspace{-5pt}
\end{equation}
Then its edge SPADE score  has the following connection with its DMD  computed using effective-resistance distances: \begin{equation}\label{edge_embed}
    \textbf{SPADE}^F(p,q) \propto \left(\gamma^F(p,q)\right)^3.
\end{equation}
\end{theorem}
The proof is available in the Appendix. So
the SPADE score of an edge $(p,q)\in E_X$ can be regarded as a surrogate for the directional derivative  $\|\nabla_{v} F(x)\|$ under the manifold setting, where $v=\pm(x_p- x_q)$. If an edge  has a larger SPADE score, it is considered  more non-robust and can be more vulnerable to   attacks along the directions formed by its end nodes.

\begin{definition}
The  \textbf{node SPADE score} is defined for  any node (data sample)  $p\in V$ as follows:
\vspace{-4pt}
\begin{equation}\label{formula_spade2}
\textbf{SPADE}^F(p)\overset{\mathrm{def}}{=}\frac{1}{|\mathbb{N}_X(p)|}\sum_{q_i\in \mathbb{N}_X(p)}^{}{\textbf{SPADE}^F(p,q_i)},
\vspace{-4pt}
\end{equation}
where $q_i\in \mathbb{N}_X(p)$ denotes the $i$-th neighbor of node $p$ in graph $G_X$, and $\mathbb{N}_X(p)\in V$ denotes the node set including all the neighbors  of $p$. 
\end{definition}

The SPADE score of a node (data sample) $p$ can be regarded as a surrogate for the function gradient $\|\nabla F(x)\|$   where  $x$ is near $p$  under the manifold setting. A node  with a larger SPADE score implies it is likely more vulnerable to adversarial attacks. 

\vspace{-5pt}
\section{Applications of  SPADE Scores}
\label{sec:app}
\textbf{Model SPADE score.} Since the SPADE score of an ML model can be used as a surrogate for the best Lipschitz constant,  we can directly use it for quantifying  the model's robustness. A greater SPADE score implies a more vulnerable ML model that can be more easily compromised with adversarial attacks. In practical applications, as long as the input and output data vectors (e.g., $X$ and $Y$) are available, the model SPADE score can be efficiently obtained by constructing the input and output graph-based manifolds (e.g., $G_X$ and $G_Y$) and subsequently computing the largest generalized eigenvalues using (\ref{formula_spade}). The detailed results are available in Section \ref{spade_metric_eval}.

\textbf{Node SPADE score.} Once we compute the node SPADE score for each input data sample as elaborated in Section \ref{sec:data_metric}, we can rank all the data samples based on their robustness scores and thus identify the most vulnerable ones, which may benefit the following applications.

 \subsection{SPADE-Guided Adversarial Training}
 A recent study shows the following findings \cite{allen2020feature}: (1) Training neural networks over the original data is   non-robust to small adversarial perturbations, while  adversarial training  can  be provably robust against any small norm-bounded perturbations; (2) The key to improving adversarial robustness is to purify non-robust features that are vulnerable to small, adversarial perturbations along the “dense mixture” directions,  via  adversarial  training; (3)  Clean training over the original data will   discover a majority of the robust features, while the adversarial training only tries to “purify” some small part of each original feature. 
 
 In practice,  some data samples may carry greater portions of non-robust features than others, which therefore should be given more attention during adversarial training. To this end, we propose an adaptive, robustness-guided adversarial training scheme  by leveraging the SPADE score of each data sample. Specifically, we use a relatively large size of the norm-bounded perturbation (epsilon ball) for data samples with top $k$ highest SPADE score, where $k$ is a hyperparameter. The details about the size of the epsilon ball and $k$ value used in our experiments are available in Section \ref{spade_adv_train}. 
 This way, much stronger defenses (with large perturbations) towards adversarial attacks should be considered only for the vulnerable data samples with the highest SPADE scores, while normal defenses (with small perturbations) will suffice for the samples with relatively small SPADE scores. The adversarial training results are available in Section \ref{spade_adv_train}.

 \subsection{SPADE-Guided Robustness Evaluation}
 \label{spade_guide_eval}
 It is worth noting that this application is orthogonal to directly applying the model SPADE score for robustness evaluation. In the latter case, as shown in Theorem \ref{l_bound}, the SPADE score is an upper bound of the smallest Lipschitz constant, which is a standalone evaluation metric. In contrast, our goal in this application is to leverage the node SPADE score to identify the most vulnerable data samples, which can guide other metric approaches and facilitate their robustness evaluation of machine learning models. For instance, to evaluate the robustness of a deep neural network using the recent CLEVER method \cite{weng2018evaluating}, a large number of data samples will be randomly selected from the original dataset; then each data sample will be processed for the CLEVER score calculation that may involve many expensive gradient computations. 
 With the guidance of node SPADE score, we only need to check a few of the most non-robust data samples to obtain a reliable CLEVER score, which will greatly improve the computation efficiency when compared with the standard practice. The results of the SPADE-guided CLEVER score are available in Section \ref{spade_clever}. 

\section{Experimental Results}
We conduct four different types of experiments to evaluate the efficacy of our proposed approach. Note that Sections \ref{spade_metric_eval} and \ref{spade_clever} exploit the SPADE metric in two orthogonal ways, as explained in Section \ref{spade_guide_eval}.

\subsection{Experimental Setup}
We obtain the input and output data used in kNN graph construction as shown below. \\
\noindent $\bullet$ \textbf{MNIST} consists of 70,000 images with the size of $28 \times 28$. We reshape each image into a $784$ dimensional vector as an input data sample. In addition, we perform inference once on a given pre-trained ML model and extract the $10$ dimensional vector right before the $softmax$ layer per image as the output data sample. Consequently, the input data $X \in {\mathbb{R} ^{70,000 \times 784}}$ and output data $Y \in {\mathbb{R} ^{70,000 \times 10}}$ are used to construct the input graph $G_X$ and output graph $G_Y$, respectively. 
  For the kNN graph construction, we choose $k=10$ ($10\sim20$) for the training (testing) set. 
  
\noindent $\bullet$ \textbf{CIFAR-10} consists of 60,000 images with the size of $32 \times 32 \times 3$. We reshape each image into a $3,072$ dimensional vector as an input data sample. Similar to MNIST, we also extract the $10$ dimensional vector right before the $softmax$ layer per image as the output data sample. Subsequently, the input data $X \in {\mathbb{R} ^{60,000 \times 3,072}}$ and output data $Y \in {\mathbb{R} ^{60,000 \times 10}}$ are used to construct input graph $G_X$ and output graph $G_Y$, respectively.
  We choose $k=100$ ($10\sim20$) for the training (testing) set in our experiments when constructing the kNN graphs.

\subsection{The SPADE Metric for Robustness Evaluation}
\label{spade_metric_eval}
\paragraph{Model SPADE scores.} To evaluate the model SPADE score as a black-box metric for quantifying model adversarial robustness, for the MNIST and CIFAR-10 test sets we compute the SPADE scores for various models trained with  different robustness levels \cite{madry2018towards}. For the MNIST dataset, $\epsilon=0.0$ to $0.3$ is considered.  For the CIFAR-10 dataset only $\epsilon=0.0$ to $2$ is considered, since for $\epsilon=4$ or $8$  the 10NN/20NN output graphs are not connected. 
 As shown in Table \ref{tab_modelspade}, the proposed model SPADE scores consistently reflect the actual levels of model adversarial robustness: in all cases the model SPADE score decreases with the increasing adversarial robustness levels.
 
\paragraph{Results of DMDs.}Table \ref{tab_edgespade} shows the average DMD values of $100$ edges selected from the input graph $G_X$. Here the geodesic distance metric is used for DMD computations, meaning that the DMD value $\gamma^F(p,q)$ of each edge $(p,q)$ in $G_X$ corresponds to the shortest-path distance between nodes $p$ and $q$ in the output graph $G_Y$. The DMD results obtained by selecting the  $100$ edges with the largest edge SPADE scores (computed using a single dominant generalized eigenvector) are labeled with ``SPADE", which are compared against the results (labeled with ``RANDOM") of the $100$ edges selected randomly from $G_X$. We make the following observations: (1) The edges selected with top SPADE scores have much greater DMDs than those selected randomly, which indicates that SPADE indeed reveals more non-robust node pairs. (2) Another expected yet noteworthy result is that for most cases, the average DMD of randomly selected edges consistently decreases with increasing adversarial robustness levels. This is because  more robust models will have a smaller model SPADE scores (upper bound of the best Lipschitz constant) and thus avoid mapping nearby data samples to distant ones. (3) The deeper models (e.g., CIFAR10 models are much deeper than the MNIST ones) map nearby data samples to more distant ones.

   \vspace{-10pt}

\begin{table}[ht!]
\caption{Model SPADE scores for  MNIST ($\epsilon = 0.0/0.1/0.2/0.3$) and CIFAR10 ($\epsilon = 0.0/0.25/0.5/1.0/2.0$). }
\label{tab_modelspade}
\begin{center}
\begin{small}
  \small	

\begin{sc}
\begin{tabular}{lccccc}
\toprule
Data Set & SPADE   (10NN) & SPADE   (20NN) \\
\midrule
MNIST                   & $42/40/37/33$ & $41/39/36/30$  \\
CIFAR10      & $432/256/200/171/79$ & $344/195/160/128/61$ \\
\bottomrule
\end{tabular}
\end{sc}
\end{small}
\end{center}
  \vspace{-20pt}
\end{table}

\begin{table}[ht!]
\caption{Average DMDs of $100$ edges in $G_X$ for the MNIST ($\epsilon = 0.0/0.1/0.2/0.3$) and CIFAR-10 ($\epsilon = 0.0/0.25/0.5/1.0/2.0$) data sets. Best results are highlighted.}
\label{tab_edgespade}
 \vspace{-5pt}
\begin{center}
\begin{small}
  \scriptsize	

\begin{sc}
\begin{tabular}{lccccc}
\toprule
Test Cases & Avg. DMD (10NN) & Avg. DMD (20NN) \\
\midrule
MNIST (SPADE)                  & $\textbf{6.6/6.1/6.5/7.7}$ & $\textbf{5.1/5.3/5.6/6.4}$  \\
MNIST (Random)                   & $3.1/2.6/2.5/2.7$ & $2.4/2.3/2.2/2.2$  \\
CIFAR10 (SPADE)      & $\textbf{11.4/11.5/9.7/12.5/8.0}$ & $\textbf{8.9/9.0/8.0/10.5/6.8}$ \\
CIFAR10 (Random)       & $6.8/6.4/5.6/5.5/5.3$ & $5.7/5.0/4.4/4.3/4.3$ \\
\bottomrule
\end{tabular}
\end{sc}
\end{small}
\end{center}
  \vspace{-15pt}
\end{table}

\subsection{SPADE-Guided Adversarial Training}
\label{spade_adv_train}
\begin{table*}[ht!]
\caption{Classification accuracy under $L_{\infty}$ bounded attacks on MNIST with $\epsilon = 0.4$.}
\label{mnist_adv}
\vspace{0pt}
\begin{center}
\begin{small}
\begin{sc}
\begin{tabular}{lccccc}
\toprule
Training Methods & Clean & PGD-50 & PGD-100 & 20PGD-100 \\
\midrule
Vanilla PGD ($\epsilon = 0.4$)      & $94.38$ & $76.89$ & $71.45$ & $70.23$ \\
PGD-Random ($\epsilon = 0.2$\&$0.4$)   & $\mathbf{97.81}$ & $87.89$ & $84.22$ & $83.09$ \\
PGD-SPADE ($\epsilon = 0.2$\&$0.4$) & $97.28$  & $\mathbf{91.65}$ & $\mathbf{89.37}$ & $\mathbf{88.46}$ \\

\bottomrule
\end{tabular}
\end{sc}
\end{small}
\end{center}
\vspace{-15pt}
\end{table*}
\begin{table*}[ht!]
   \vspace{-0pt}
\caption{Classification accuracy under $L_{\infty}$ bounded attacks on CIFAR-10 with $\epsilon = 2.0/4.0/8.0$.}
\label{cifar_adv}
\begin{center}
\begin{sc}
\begin{tabular}{lccccc}
\toprule
Training Methods & Clean & 10PGD-50 \\
\midrule
Vanilla PGD ($\epsilon = 8.0$)      & $\mathbf{81.57}$ & $75.28/67.92/50.35$ \\
Vanilla PGD ($\epsilon = 12.0$)      & $76.93$ & $71.41/65.58/51.62$ \\
Vanilla PGD ($\epsilon = 14.0$)      & $75.12$ & $70.12/64.64/51.79$ \\
PGD-Random ($\epsilon = 12.0$\&$14.0$)      & $76.09$ & $70.87/65.73/51.77$ \\
PGD-SPADE ($\epsilon = 12.0$\&$14.0$)      & $81.38$ & $\mathbf{75.74/69.19/53.61}$  \\

\bottomrule
\end{tabular}
\end{sc}
\end{center}
 \vspace{-16pt}
\end{table*}
\begin{table*}[ht!]
\caption{Comparison of   CLEVER scores for robustness evaluation of ML models \cite{weng2018evaluating}. ``CNN", ``DD", and ``2CL" stand for the 7-layer AlexNet-like, Defensive Distillation and 2-convolutional-layer CNNs, respectively.  The SPADE-guided CLEVER scores  are computed using top $10$ (``T10")   non-robust samples
and  compared against the CLEVER scores computed with $10$ (``R10")  and $100$ (``R100") randomly selected samples from  the MNIST/CIFAR-10 test sets.  }
\label{clever_k10}
\begin{center}
\begin{small}
\small	
\begin{sc}
\begin{tabular}{lccccc}
\toprule
Networks & SPADE (10NN,T10) &  SPADE (20NN, T20) & CLEVER (R10)& CLEVER (R100)\\
\midrule
MNIST-MLP     & $1.317/0.067$ &$ \mathbf{0.590/0.030}$ & $0.698/0.034$ &$0.819/0.041$\\
MNIST-CNN     & $\mathbf{0.379/0.030}$ &$ 0.391/0.027$ & $0.775/0.057$ &$0.721/0.057$\\
MNIST-DD      & $\mathbf{0.408/0.026}$ &$0.451/0.028$ & $0.874/0.065$ &$0.865/0.063$\\
CIFAR-MLP     & $\mathbf{0.213/0.004}$ &$0.226/0.005$ & $0.312/0.007$ &$0.219/0.005$\\
CIFAR-CNN     & $0.141/0.004$ &$0.088/0.003$ & $\mathbf{0.046/0.001}$ &$0.072/0.002$\\
CIFAR-DD      & $0.310/0.009$ &$ 0.119/0.003$ & $\mathbf{0.100/0.003}$ &$0.130/0.004$\\
MNIST-2CL($\epsilon = 0$)     & $\mathbf{0.049/0.002}$  &$0.075/ 0.003$ & $162.35/7.592$ &$68.544/3.182$\\
MNIST-2CL($\epsilon = 0.3$)   & $\mathbf{0.112/0.008}$ &$0.114/0.006$ & $0.332/0.017$ &$0.431/0.022$\\

\bottomrule
\end{tabular}
\end{sc}
\end{small}
\end{center}
\vspace{-10pt}
\end{table*}

We choose LeNet-5 and ResNet-18 as basic CNN models on MNIST and CIFAR-10, respectively~\cite{lecun2015lenet,he2016deep}. Moreover, we evaluate several baselines as well as our method on MNIST and CIFAR-10 shown below:\\
  \noindent $\bullet$ \textbf{Vanilla PGD.} The vanilla projected gradient decent (PGD) based adversarial training approach with perturbation magnitude $\epsilon \in \{0.4\}$ and $\{8.0, 12.0, 14.0\}$ on MNIST and CIFAR-10, respectively. \cite{madry2018towards}.\\
  \noindent $\bullet$ \textbf{PGD-Random.} The PGD-based training method but randomly pick $\epsilon$ from $\{0.2, 0.4\}$ ($\{12.0, 14.0\}$) for each training image on MNIST (CIFAR-10).\\
  \noindent $\bullet$\textbf{PGD-SPADE (Our method).} The PGD-based training method with $\epsilon = 0.3$ ($14.0$) for top 45,000 non-robust images guided by the node SPADE scores, and $\epsilon=0.3$ ($12.0$) for rest of images on MNIST (CIFAR-10). In addition, to enhance the clean accuracy on CIFAR-10, we skip performing PGD on images that are misclassfied without adversarial perturbation, as suggested in~\cite{balaji2019instance, cheng2020cat}. \\

\textbf{MNIST.} We report the averaged classification accuracy over $8$ runs on clean test images as well as perturbed images under $3$ different $L_{\infty}$ bounded attacks with $\epsilon = 0.4$: PGD attack with the PGD iteration of 50 (PGD-50) and 100 (PGD-100), and PGD-100 attack with 20 random restarts (20PGD-100)~\cite{madry2018towards}. All the attacks use $0.01$ step size. As shown in Table \ref{mnist_adv}, our method achieves at least $5.37\%$ accuracy improvement compared against all baselines under the strongest attack (i.e., 20PGD-100). It is worth noting that PGD-SPADE consistently improves the accuracy over PGD-Random under different attacks, which indicates that SPADE can identify the robust images and choose variable epsilon balls accordingly to improve the adversarial accuracy.

\textbf{CIFAR-10.} We report the averaged classification accuracy over $8$ runs on clean as well as perturbed test images under a strong $L_{\infty}$ bounded attack with $\epsilon \in \{2.0, 4.0, 8.0\}$: PGD attack with 10 random restarts and 50 iterations (i.e., 10PGD-50). All the attacks use a step size of $2.0$. Table \ref{cifar_adv} shows that our SPADE-guided PGD training method improves the accuracy of PGD-Random as well as the vanilla PGD with all different perturbation magnitudes. This reveals that training with variable epsilon balls guided by the SPADE score indeed enhances the model robustness.

\subsection{SPADE-Guided Robustness Evaluation}
\label{spade_clever}


We choose MNIST and CIFAR-10  for robustness evaluation based on the CLEVER method \cite{weng2018evaluating}. For both datasets, we evaluate CLEVER scores on three networks: a single hidden layer multilayer perceptron (MLP) with the default number of hidden units \cite{hein2017formal}, a $7$-layer AlexNet-like CNN with the same structure described in \cite{CW}, a $7$-layer CNN with defensive distillation \cite{papernot2017practical}. For MNIST, we also evaluate CLEVER scores on two $2$-convolutional-layer CNNs \cite{madry2018towards} trained with  different robustness levels ($\epsilon=0.0$ and $\epsilon=0.3$). For comparison, we  compute  the SPADE-guided CLEVER scores for the same datasets using the same networks. 


\begin{figure}[ht]
\centering
  \subfigure{\includegraphics[width=0.99\columnwidth]{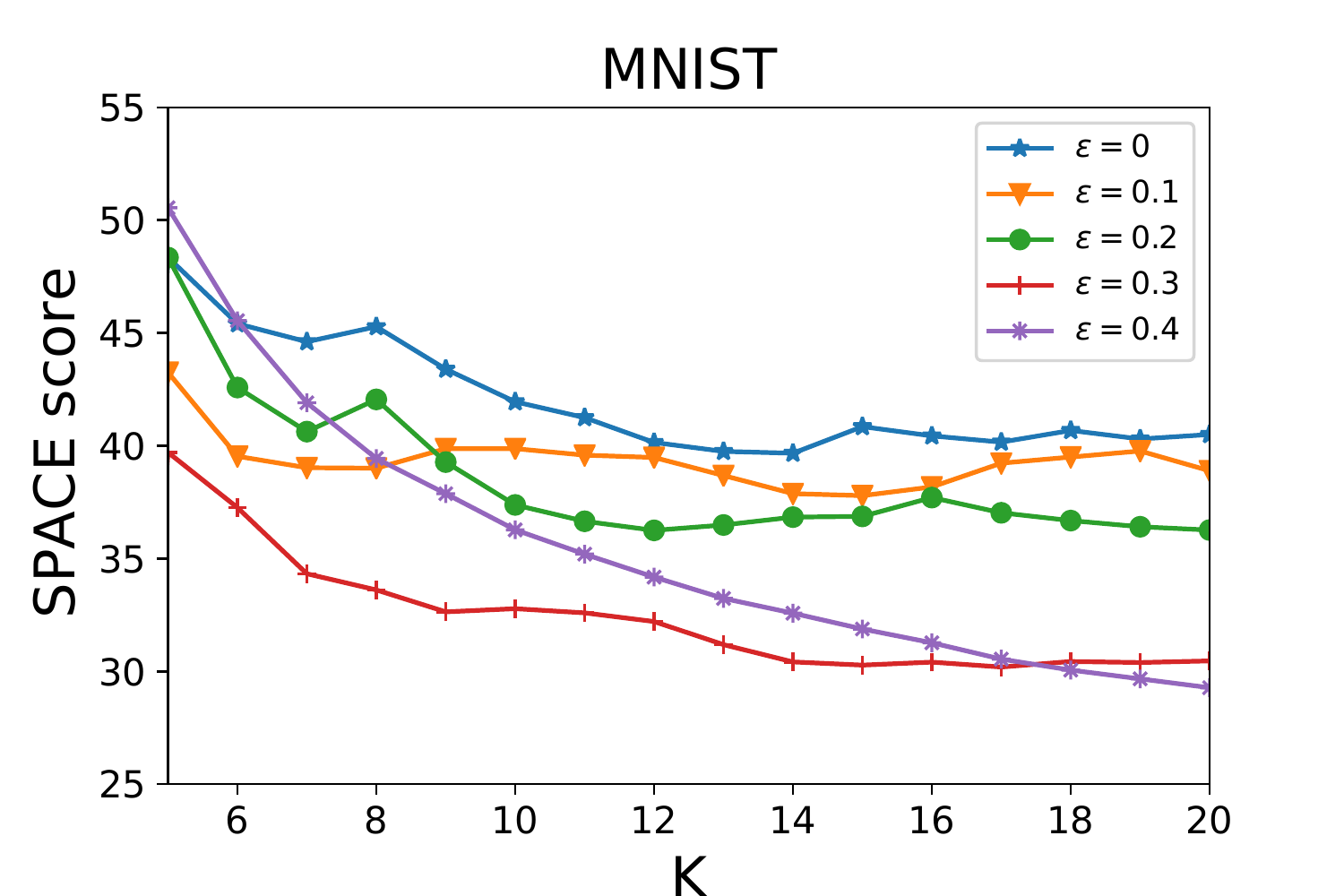}}
  \subfigure{\includegraphics[width=.9975\columnwidth]{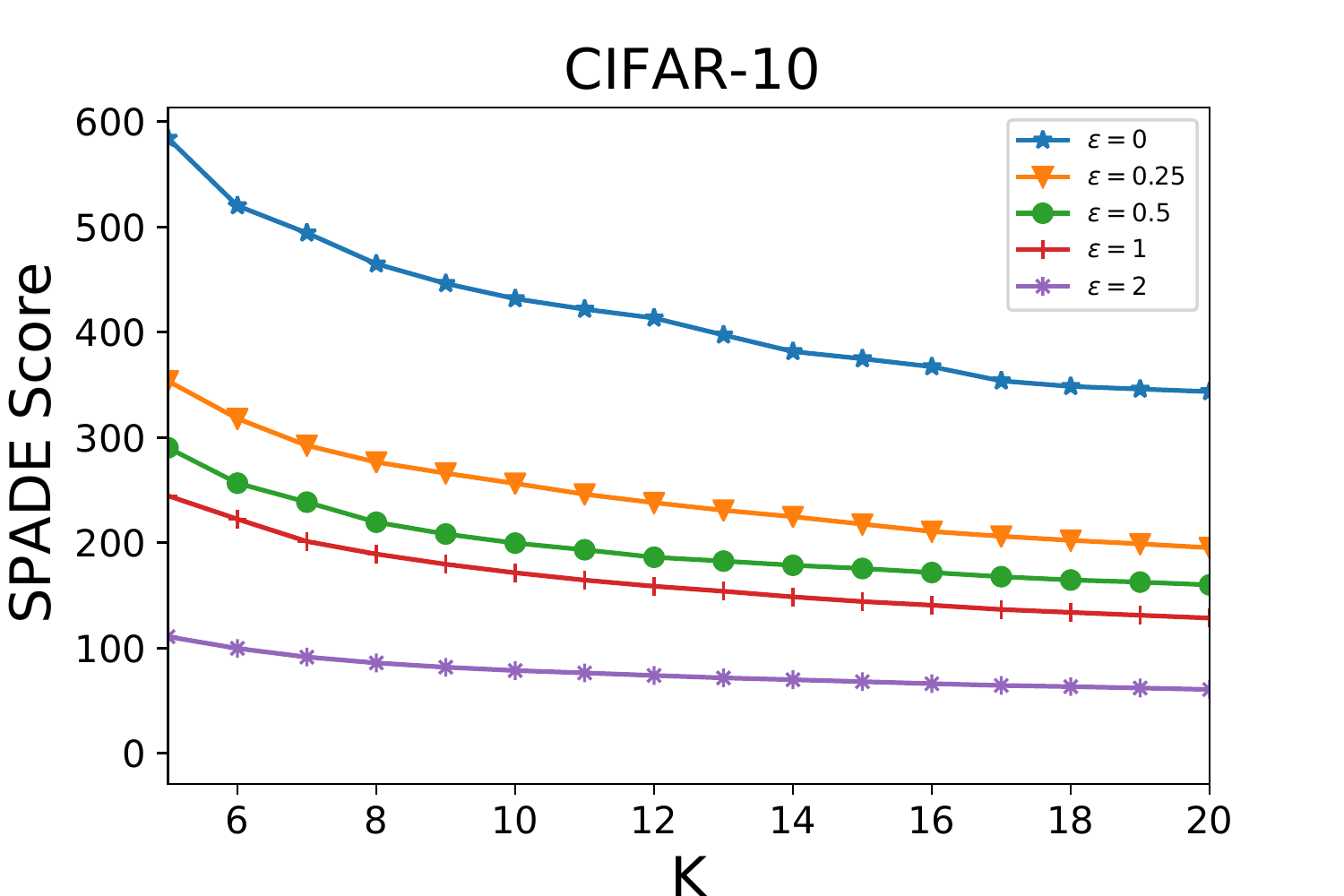}}
  \vspace{-10pt}
\caption{Model SPADE score with varying k values of kNN graph.}
\label{fig:ablation}
\end{figure}
For the aforementioned experiments, we use the default sampling parameters in \cite{weng2018evaluating}. Since computing CLEVER score for each image sample can already be time consuming, we only choose $10$ test-set image samples for conducting the untargeted attacks for both MNIST and CIFAR-10. We show the experiment results in Table \ref{clever_k10}. As expected,  in most cases our SPADE-guided CLEVER scores are much smaller than the normal CLEVER scores computed based on randomly selected samples.  We also observe that when evaluating CLEVER scores based on a relatively small sample set, the results can be significantly biased. For instance, in the case for MNIST-2CL ($\epsilon = 0$), our SPADE-guided score is over $3,000\times$ smaller than the original. Consequently, directly applying CLEVER evaluations without the guidance of SPADE scores may not help correctly assess the model  robustness. Here only the SPADE-guided CLEVER scores show consistent robustness evaluations for the MNIST-2CL models trained under $\epsilon=0.0$ and $\epsilon=0.3$ settings.

\subsection{Ablation Study on k Value of kNN Graphs}
To study the sensitivity of SPADE score to the k value of kNN graph, we use the SPADE score to evaluate non-robustness of models with different k for constructing the kNN graphs. Specifically, we use adversarially trained LeNet5 (ResNet50) model with $\epsilon \in \{0, 0.1, 0.2, 0.3, 0.4\}$ ($\epsilon \in \{0, 0.25, 0.5, 1, 2\}$) on MNIST (CIFAR-10). The model with  higher $\epsilon$  is  more adversarially robust. We further vary k from $5$ to $20$ for constructing  kNN graphs. As shown in Figure \ref{fig:ablation}, the SPADE scores consistently reveal the model non-robustness on CIFAR-10. For the results on MNIST, SPADE score fails to reflect the model non-robustness (e.g., the model with $\epsilon=0.4$) when k is too small ($k<10$). However, the SPADE score gradually captures model non-robustness when increasing the k value and correctly ranks the non-robustness of all models with $k=20$. Thus, a relatively large k (e.g., $20$) is preferred when constructing kNN graphs for computing the SPADE score.


\section{Conclusion}

This work introduces SPADE, a black-box spectral method for evaluating the adversarial robustness of a given ML model based on graph-based manifolds. We formally prove that the proposed SPADE metric is an upper bound of the best Lipschitz constant under the manifold setting. Moreover, we extend the SPADE score to identify the most non-robust data samples that are potentially vulnerable to adversarial perturbations. Our extensive experiments show that the model SPADE score is a good surrogate for the best Lipschitz constant, and thus can be leveraged for revealing the  level of adversarial robustness of a given  ML model. In addition, our results show that   the sample SPADE scores can be exploited for enhancing the performance of existing adversarial training as well as  adversarial robustness evaluations.

\section{Acknowledgments}
This work is supported in part by the National Science Foundation under Grants  CCF-2041519,  CCF-2021309, CCF-2011412, and CCF-2007832. The authors would like to thank Weizhe Hua (Cornell) and Shusen Wang (Stevens) for their helpful discussions on the topic of adversarial training. 
\bibliography{spade}
\bibliographystyle{icml2021}

\clearpage

\appendix

\section{Supplementary Materials}

\subsection{Proof for Theorem $\mathbf{1}$}
\label{proof_t1}

\begin{proof}
We first compute effective-resistance distance between nodes $p$ and $q$ on  $G_Y$ and $G_X$ by $d_Y(p,q)=e_{p,q}^\top L_Y^+ e_{p,q}$ and $d_X(p,q)=e_{p,q}^\top L_X^+ e_{p,q}$, respectively. Consequently, $\gamma^F_{max}$ satisfies the following inequality:
\begin{equation}\label{formula_eigbound2}
\gamma^F_{max}=\mathop{\max_{\forall p, q\in V}}_{p\neq q}\frac{e_{p,q}^\top L_Y^+ e_{p,q}}{e_{p,q}^\top L_X^+ e_{p,q}}\le\mathop{\max_{||{v}||\neq 0}}_{v^\top \mathbf{1}=0 }\frac{v^\top L_Y^+ v}{v^\top L_X^+ v}
\end{equation}
where $\mathbf{1}\in {\mathbb{R} ^{N}}$ denotes the all-one vector. According to the generalized {Courant-Fischer theorem,}   
 we have:
\begin{equation}\label{formula_eigbound3}
\mathop{\max_{||{v}||\neq 0}}_{v^\top \mathbf{1}=0 }\frac{v^\top L_Y^+ v}{v^\top L_X^+ v}=\lambda_{max}(L_X L_Y^+)=\lambda_{max}( L_Y^+L_X )
\end{equation}
By combining Equations \eqref{formula_eigbound2} and \eqref{formula_eigbound3}, we have:
\begin{equation}\label{formula_eigbound4}
\gamma^F_{max}\le\mathop{\max_{||{v}||\neq 0}}_{v^\top \mathbf{1}=0 }\frac{v^\top L_X v}{v^\top L_Y v}= \lambda_{max}( L_Y^+L_X )
\end{equation}
which completes the proof of the theorem.
\end{proof}


\subsection{Proof for Theorem $\mathbf{2}$}
\label{proof_t2}

\begin{proof} Before proving our theorem, we first introduce the formal definition of node subset and its boundary in a graph:
\begin{definition}\label{def_nodeset}
Given a node coloring vector $z\in {\mathbb{R}^N}$  including only $0$s and $1$s,  a  \textbf{node subset} $S_z \subset V$ and its complement   $ \overline{S}_z \subset V$ are defined as follows, respectively:
\begin{equation}
\begin{split}
    S_z \overset{\mathrm{def}}{=}\left\{ p \in V: z(p)=1 \right\},\\
    \overline{S}_z  \overset{\mathrm{def}}{=}\left\{ p \in V: z(p)=0 \right\}.
\end{split}
\end{equation}
\end{definition}

\begin{definition}
The \textbf{boundaries} $\partial_{G_X} (S_z)$ and $\partial_{G_Y} (S_z)$  of $S_z$ in graphs $G_X$ and $G_Y$  are defined as follows, respectively:
\begin{equation}
\begin{split}
\partial_{G_X} (S_z) \overset{\mathrm{def}}{=}\left\{ (p,q)\in E_X: p\notin  S_z, q \in S_z\right\},\\
\partial_{G_Y} (S_z) \overset{\mathrm{def}}{=}\left\{ (p,q)\in E_Y: p\notin  S_z, q \in S_z\right\}.
\end{split}
\end{equation}
\end{definition}

Consequently, the \textbf{cut}  (the size of the boundary) of  $S_z$  in each of  $G_X$ and $G_Y$ can be computed as follows, respectively:
\begin{equation}
\begin{split}
cut_X(S_z,  \overline{S}_z  )&={z}^\top {L}_X {z}= |\partial_{G_X} (S_z)|, \\
cut_Y(S_z,  \overline{S}_z  )&={z}^\top {L}_Y {z}= |\partial_{G_Y} (S_z)|. 
\end{split}
\end{equation}

According to the generalized {Courant-Fischer theorem,} 
  we have the following inequality:
\begin{align*}
  \lambda_{max}( L_Y^+L_X ) &= \mathop{\max_{|{v}|\neq 0}}_{v^\top \mathbf{1}=0 }\frac{v^\top L_X v}{v^\top L_Y v} \\
  &\ge \mathop{\max_{\forall S_z\subset V}}_{}\frac{{z}^\top {L}_X {z}}{{z}^\top {L}_Y {z}} \\
  &= \mathop{\max_{\forall S_z\subset V}}\frac{|\partial_{G_X} (S_z)|}{|\partial_{G_Y} (S_z)|} \\
  &= \frac{1}{\zeta^F_{min}}
\end{align*}
which completes the proof of the theorem.
\end{proof}

\subsection{Proof for Theorem $\mathbf{3}$}
\label{proof_t3}
\begin{proof}
Let $u_1, u_2, ..., u_N$ and $v_1, v_2, ..., v_N$ denote the $N$  eigenvectors of $L_X L_Y^+$ and $L_Y^+L_X$, respectively, while their corresponding  shared eigenvalues are denoted by $\lambda_1, \lambda_2, ..., \lambda_N$.  In addition, eigenvectors ${{u_i}}$ can  be constructed to satisfy:
\begin{equation}\label{formula_p-orth}
{u_i^\top L^+_X u_j^{}} = \left\{ \begin{array}{l}
1, ~~~i = j\\
0, ~~~i \ne j.
\end{array} \right.
\end{equation}
\begin{equation}\label{formula_p-orth2}
\Rightarrow{u_i^\top L^+_Y u_j^{}} = \left\{ \begin{array}{l}
\lambda_i, ~~~i = j\\
0, ~~~i \ne j.
\end{array} \right.
\end{equation}
Therefore, the following equations hold:
 \begin{equation}\label{eigen_dual}
 \begin{split}
 L_Y^+u_i&=\lambda_i L_X^+ u_i  \leftrightarrow L_Y^+L_X \left(L_Y^+u_i\right)=\lambda_i \left(L_Y^+u_i\right)\\ L_X v_i&=\lambda_i L_Y v_i \leftrightarrow L_Y^+L_X v_i=\lambda_i v_i 
\end{split}
\end{equation}
which leads to the following equation
\begin{equation}\label{eigen_dual2}
 \begin{split}
   v_i&=\beta_i L_Y^+ u_i\\ \Rightarrow u^\top_j v_i  &= \left\{ \begin{array}{l}
\beta_i \lambda_i, ~~~i = j\\
0, ~~~i \ne j.
\end{array} \right. 
\end{split}
\end{equation}
where $\beta_i$ denotes a scaling coefficient.
Without loss of generality,  $e_{p,q}$ can be expressed as a linear combination of  $u_i$ for $i=1,...,N$ as follows:
 \begin{equation}\label{formula_epq}
 e_{p,q}=\sum_{i=1}^N \alpha_i u_i.
\end{equation}
Then
$\gamma^F(p,q)$ can be rewritten as follows:
\begin{equation} \label{gammanew}
\begin{split}
\gamma^F(p,q)&=\frac{d_Y(p,q)}{d_X(p,q)}=\frac{e_{p,q}^\top L_Y^+ e_{p,q}}{e_{p,q}^\top L_X^+ e_{p,q}}\\
&=\frac{(\sum_{i=1}^N \alpha_i u_i)^\top L_Y^+(\sum_{i=1}^N \alpha_i u_i) }{(\sum_{i=1}^N \alpha_i u_i)^\top L_X^+(\sum_{i=1}^N \alpha_i u_i)}\\
&=\frac{\sum_{i=1}^N\sum_{j=1}^N \alpha_i\alpha_j u_i^\top L_Y^+u_j }{\sum_{i=1}^N\sum_{j=1}^N \alpha_i\alpha_j u_i^\top L_X^+u_j}\\
&=\frac{\sum_{i=1}^N \alpha^2_i u_i^\top L_Y^+u_i }{\sum_{i=1}^N \alpha^2_i  u_i^\top L_X^+u_i}\\
&=\frac{\sum_{i=1}^N\alpha^2_i\lambda_i}{\sum_{i=1}^N\alpha^2_i}.
\end{split}
\end{equation}
If the edge $({p,q})$  is  dominantly aligned with a single dominant generalized eigenvector $u_k$ where $1\le k \le r$, it  implies  $\forall i\neq k, \alpha_i \approx 0$ and thus $e_{p,q}\approx \alpha_k u_k$. Then $\gamma^F(p,q)$ can be approximated by:
\begin{equation} \label{gammanew2}
\begin{split}
\gamma^F(p,q)\approx \lambda_k.
\end{split}
\end{equation}
On the other hand,  (\ref{eigen_dual2}) allows the edge SPADE score of $({p,q})$ to be expressed as
\begin{equation}
\begin{split}
    \textbf{SPADE}^F(p,q)&=\|V_r^\top e_{p,q}\|_2^2 \\
    &=\sum_{i=1}^r \lambda_i (v_i^\top e_{p,q})^2 \\
    &=\sum_{i=1}^r \lambda_i \left(\sum_{j=1}^N \alpha_j \beta_i u_j^\top L_Y^+ u_i \right)^2 \\
    &=\sum_{i=1}^r \alpha^2_i \beta^2_i\lambda^3_i  \\
    &\approx \alpha^2_k \beta^2_k \lambda^3_k \propto \left(\gamma^F(p,q)\right)^3
    \end{split}
\end{equation}
which completes the proof of the theorem.
\end{proof}

\subsection{A Spectral Perspective of Adversarial  Training}
 
{The   Riemannian distance}   between positive definite  (PSD) matrices 
has been considered as the most natural and useful distance defined on the positive
definite cone $\mathbb{S}^n_{++}$~\cite{bonnabel2010riemannian}.
\begin{definition}
The  \textbf{ Riemannian distance} $\delta({L_{X}},{L_{Y}})$ between $L_X$ and $L_Y$ is given by~\cite{lim2019geometric}:
\begin{equation}\label{formula_RD}
      \delta({L_{Y}},{L_{X}})=\delta({L_{X}},{L_{Y}}) \overset{\mathrm{def}}{=}\left[\sum_{i=1}^{N}\log^2\lambda_i(L_Y^+L_X)\right]^{\frac{1}{2}}. 
\end{equation}
\end{definition}

The Riemannian distance $\delta({L_{X}},{L_{Y}})$ requires all eigenvalues of $L_Y^+L_X$ to be evaluated, and therefore   encapsulates all possible  cut mapping distortions.

While $\lambda_{max}(L_Y^+L_X)$    tells if   function $F$ can map two nearby data points (nodes) to distant ones at output,
 $\lambda_{max}(L_X^+L_Y)$ tells if there exist  two neighboring output data points (nodes)  that are mapped from two distant ones at input.  Since $\lambda_{max}(L_Y^+L_X)\gg\lambda_{max}(L_X^+L_Y)$ holds for most machine learning applications,  $\lambda_{max}(L_Y^+L_X)$ will be more interesting and relevant to adversarial robustness.

\begin{remark}
The Riemannian distance $\delta({L_{Y}},{L_{X}})$ for a typical ML model will be upper bounded by:
\begin{equation}\begin{split}
    \delta_{}({L_{Y}},{L_{X}})\le N\log \lambda_{max}(L_Y^+L_X).
\end{split}
\end{equation} 
\end{remark}


\begin{definition} Given two metric spaces $(X, dist_X)$ and $(Y, dist_Y)$, where $dist_X$ and $dist_Y$ denote the metrics on the sets $X$ and $Y$ respectively, if there exists a $\kappa \ge 1$ with:
\begin{equation}
     \frac{1}{\kappa} dist_X(p,q)\le dist_Y(p,q)\le \kappa dist_X(p,q),
\end{equation}
then $Y=F(X)$ is called a \textbf{$\kappa$-bi-Lipschitz  mapping}, which is injective, and is   a homeomorphism onto its image.
\end{definition}
Choosing $\kappa=\lambda_{max}(L_Y^+L_X)=\textbf{SPADE}^F$ will make $Y=F(X)$ a $\kappa$-bi-Lipschitz  mapping under the manifold setting. 

 \begin{remark}
 Adversarial training can effectively decrease the Riemannian distances between the input and output manifolds, which is similar to decreasing the Lipschitz constant.
 \end{remark}
 
\begin{figure}[ht]
\vskip -0.in
\begin{center}
\centerline{\includegraphics[width=0.5\textwidth]{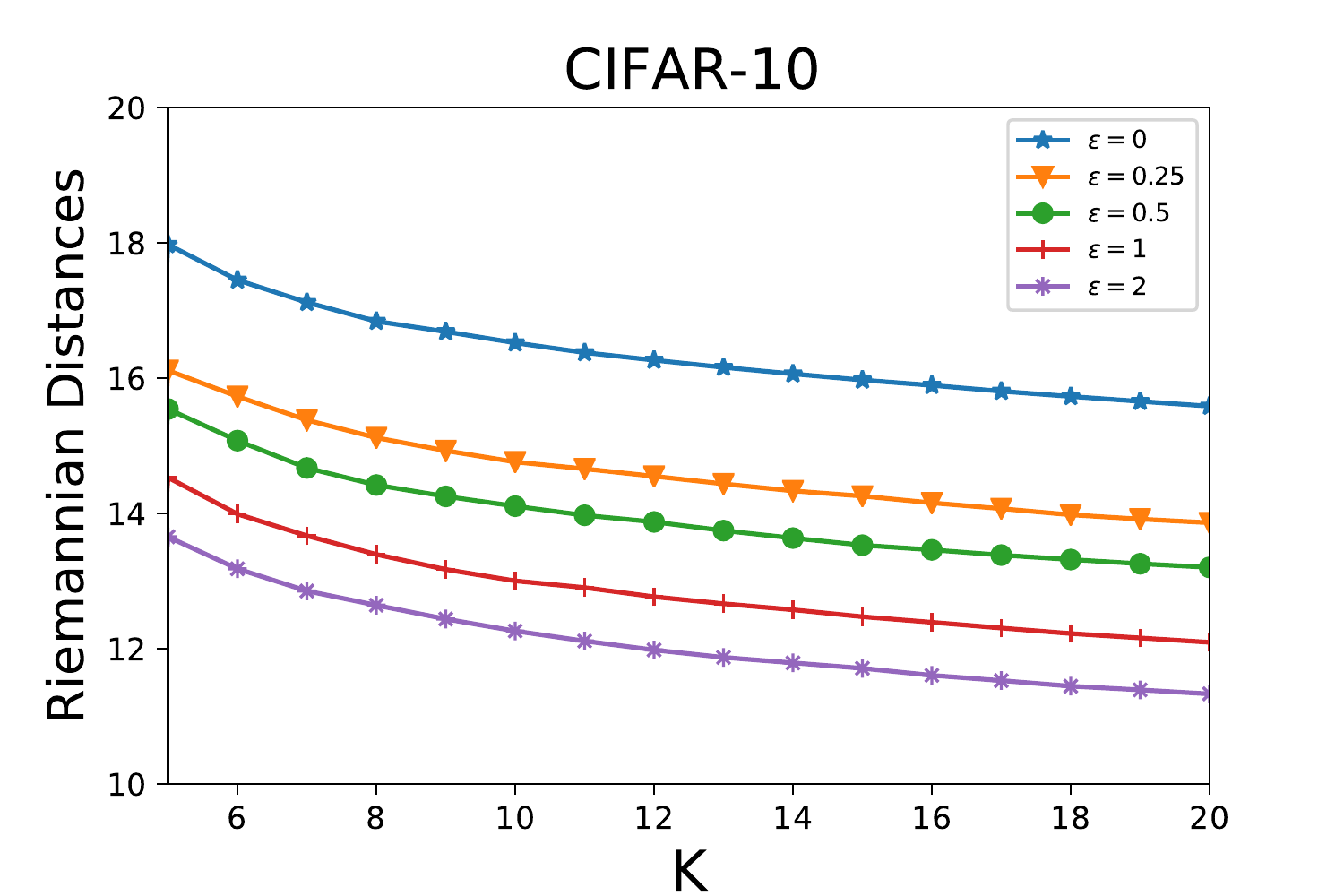}}
\caption{Riemannian distances of   adversarially-trained models. A greater $\epsilon$ value indicates a more adversarially robust NN model. }
\label{fig:RD}
\end{center}
\vskip -0.2in
\end{figure}  
\textbf{An example.} We demonstrate the Riemannian distances of the PGD trained neural networks  in  Figure \ref{fig:RD}. Our results are obtained using the CIFAR-10 data set, which includes $10,000$ images (data points). We construct $k$-nearest-neighbor (kNN) graphs for approximating the input/output data manifolds with varying $k$ values. 
The Riemannian distances have been  calculated only using the $10$ largest eigenvalues of $L_Y^+L_X$.
 As shown in Figure \ref{fig:RD},   the more adversarially robust neural networks  always  exhibit lower Riemannian distances. 
 

\subsection{Eigensolver for computing SPADE score}
\label{sec:eigensolver}
\begin{table*}[ht!]
\caption{Eigensolver performance when using \texttt{eigs} and power iteration method for MNIST and CIFAR-10 data sets with $\epsilon = 0$.}
\label{eigensolver}
\vskip 0.15in
\begin{center}
\begin{small}
\begin{sc}
\begin{tabular}{lllll}
\toprule
 & \multicolumn{2}{c}{MNIST} & \multicolumn{2}{c}{CIFAR-10} \\
 & {eigs} & power\_iteration & {eigs}  & power\_iteration\\
\midrule
 $K=10$, $\lambda_{max}(L_Y^+L_X)$ & 98.64 & 98.38 (err = 0.26\%) & 398.16 & 396.48 (err = 0.42\%) \\
$K=10$, runtime  & 9.22s & 5.88s & 11.96s & 2.65s\\
$K=20$, $\lambda_{max}(L_Y^+L_X)$  & 81.14 & 80.83 (err = 0.38\%) & 311.49 & 310.21 (err = 0.41\%)\\
$K=20$, runtime &16.19s & 7.62s & 19.40s & 7.29s\\
\bottomrule
\end{tabular}
\end{sc}
\end{small}
\end{center}
\vskip -0.1in
\end{table*}

To efficiently find the SPADE score, power iteration method can be leveraged to compute the dominant eigenvalues of $L_Y^+L_X$. To further reduce the complexity, graph-theoretic algebraic multigrid solvers~\cite{koutis2011combinatorial, livne2012lean} 
can be applied to solve the corresponding graph Laplacians. 
To evaluate the performance of the power iteration method, we calculated the largest eigenvalue of $L_Y^+L_X$ using built-in MATLAB \texttt{eigs} function and the power iteration method, where $L_X$ and $L_Y$ represent the kNN graph Laplacians constructed based on MNIST and CIFAR-10 dataset with K being 10 and 20. As shown in Table \ref{eigensolver}, $\lambda_{max}(L_Y^+L_X)$ and the corresponding run time are reported in the table under different settings; ERR represents the relative error of $\lambda_{max}(L_Y^+L_X)$ comparing with \texttt{eigs}. From the table we  observe that power iteration method allows efficiently computing   $\lambda_{max}(L_Y^+L_X)$ with satisfactory accuracy levels.

\subsection{Spectral Embedding and SPADE Scores}
\begin{figure*}[ht]
\vskip 0.2in
\begin{center}
\centerline{\includegraphics[width=2\columnwidth]{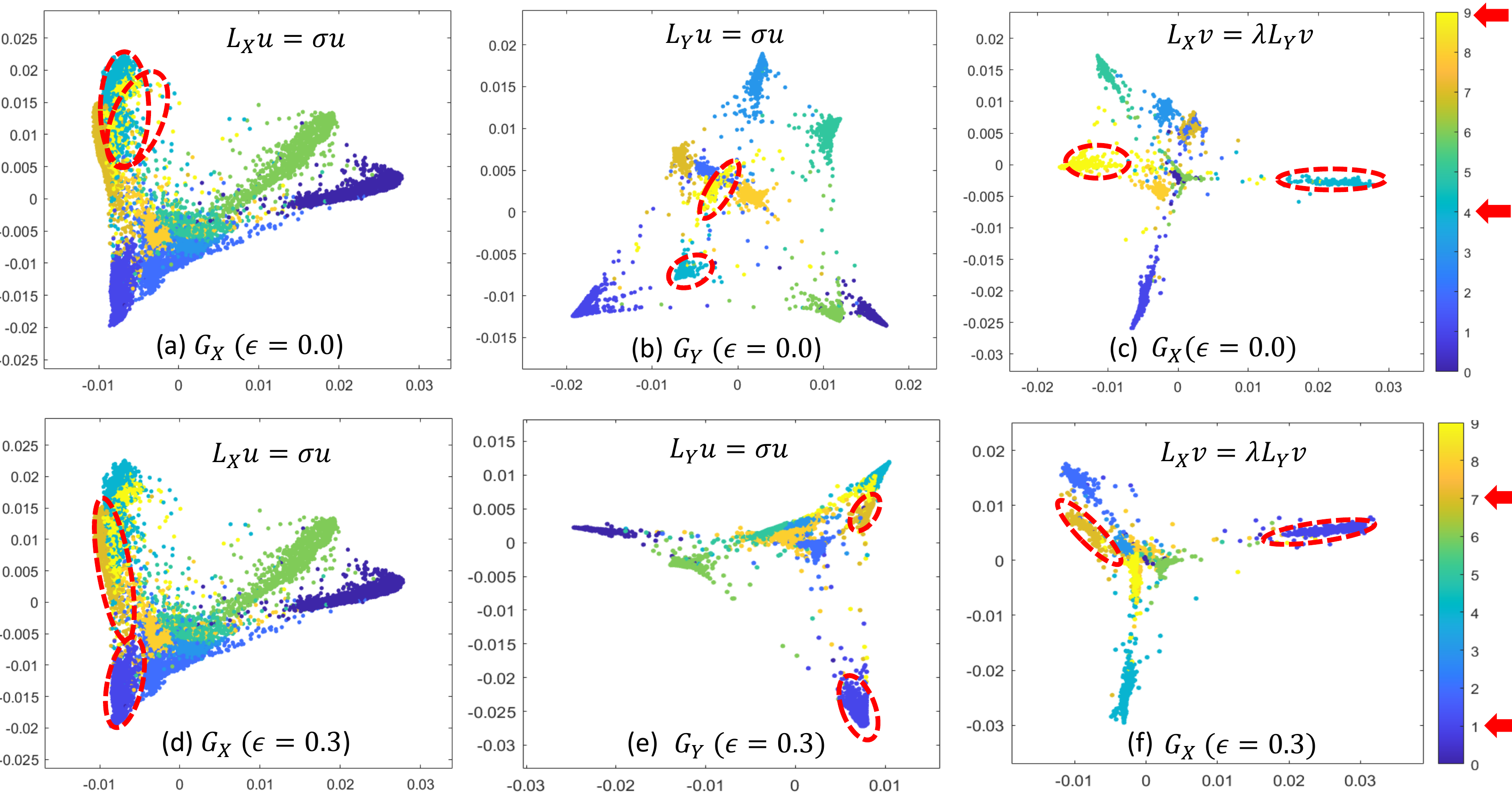}}
\caption{Revealing non-robust node pairs via spectral embedding w/ the first two Laplacian  and generalized eigenvectors before (top row) and after (bottom row) adversarial  training. The labels of the most non-robust node pairs are indicated by red arrows for both settings.}
\label{fig:MNISTdual}
\end{center}
\vskip -0.2in
\end{figure*}

Figure \ref{fig:MNISTdual} shows the 2D spectral node  embeddings  before/after adversarial training with the first two eigenvectors of $L_X$, $L_Y$ and $L^+_YL_X$, respectively, for the   MNIST test   set that includes $10,000$ handwritten digits from $0$ to $9$. The edges are not shown for the sake of clarity. The following has been observed: \textbf{(1)} {Before  adversarial training}, the handwritten digits of   $4$ and $9$ are very close at input but much more separated at output as shown in Figures (a) and (b),   which implies a rather poor adversarial robustness level; a similar conclusion can be made by checking the node embedding   with generalized eigenvectors   in Figure (c), where $4$ and $9$ clusters are most separated from each other. \textbf{(2)} {After adversarial training},   the $4$ and $9$ clusters  are much closer at output as shown in Figure  (e), which is due to the  substantially improved robustness, while digits $1$ and $7$  become the most non-robust  ones as shown in Figure (f). 
\begin{figure*}[ht]
\vskip 0.2in
\begin{center}
\centerline{\includegraphics[width=0.95\columnwidth]{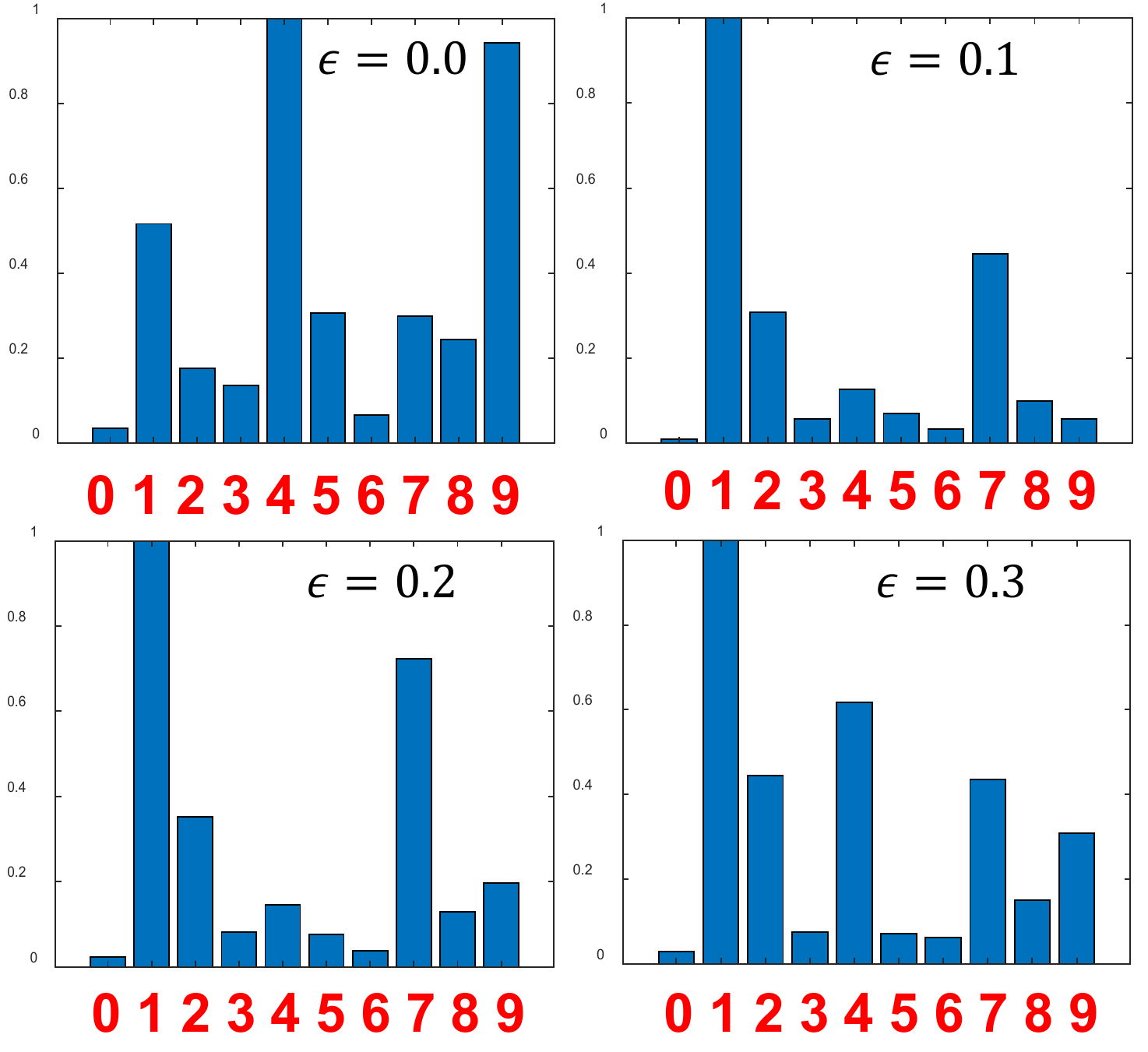}\includegraphics[width=0.95\columnwidth]{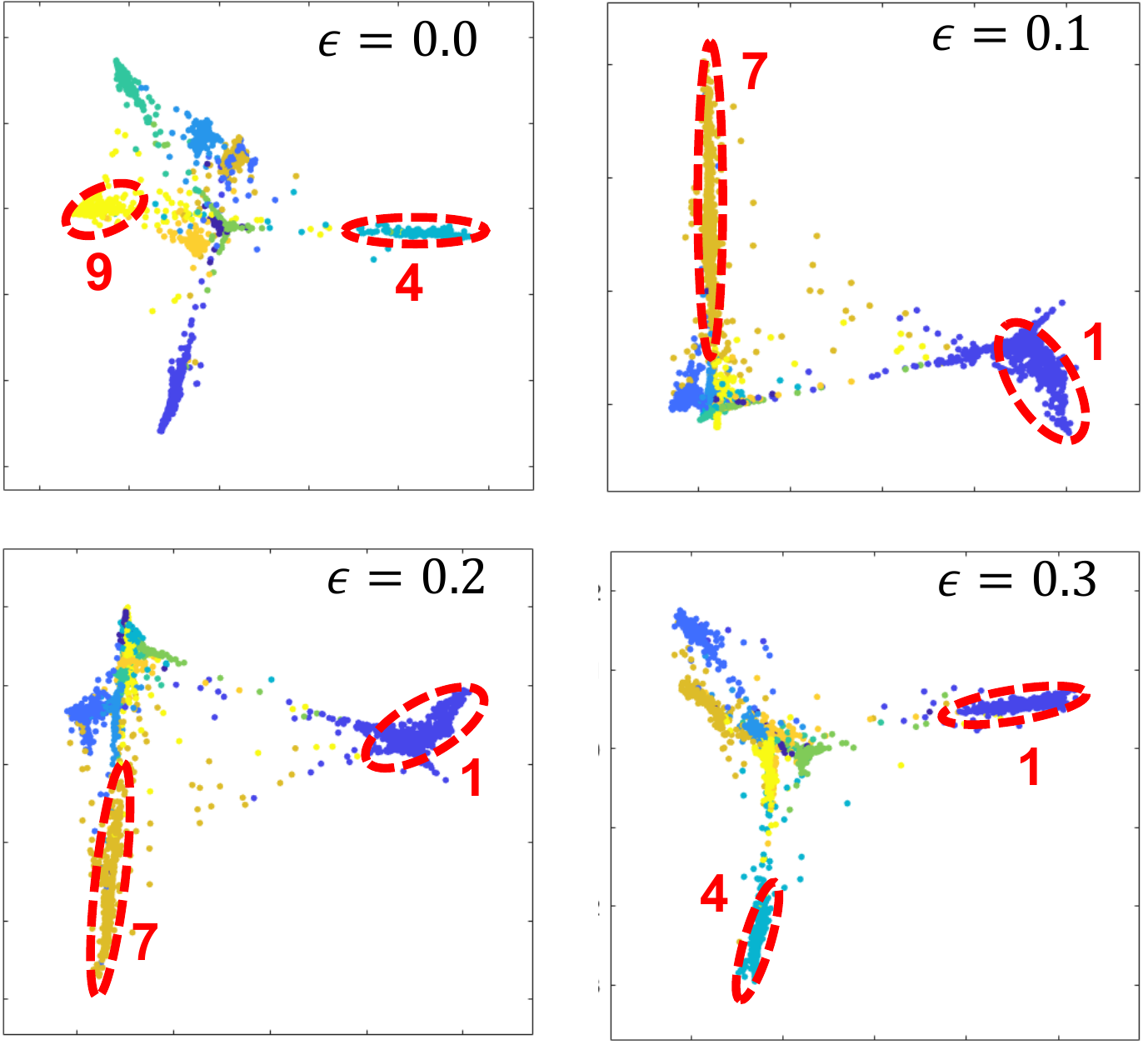}}
\caption{The label SPADE scores and spectral   embeddings   with top two dominant generalized eigenvectors (MNIST test set). }
\label{fig:spadeMNIST}
\end{center}
\vskip -0.2in
\end{figure*}

\begin{figure*}[ht]
\vskip -0.1in
\begin{center}
\centerline{\includegraphics[width=0.95\columnwidth]{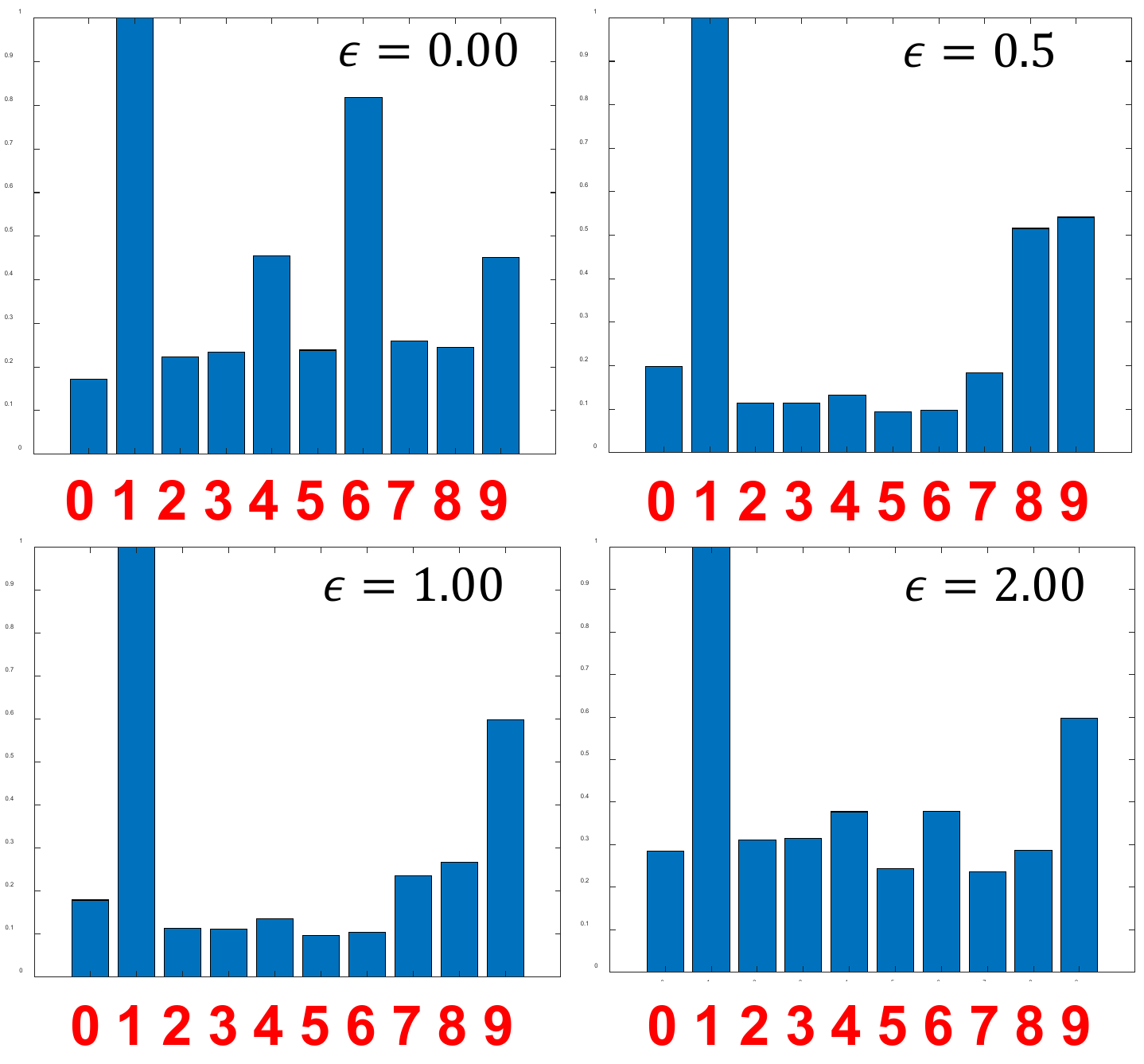}\includegraphics[width=.985\columnwidth]{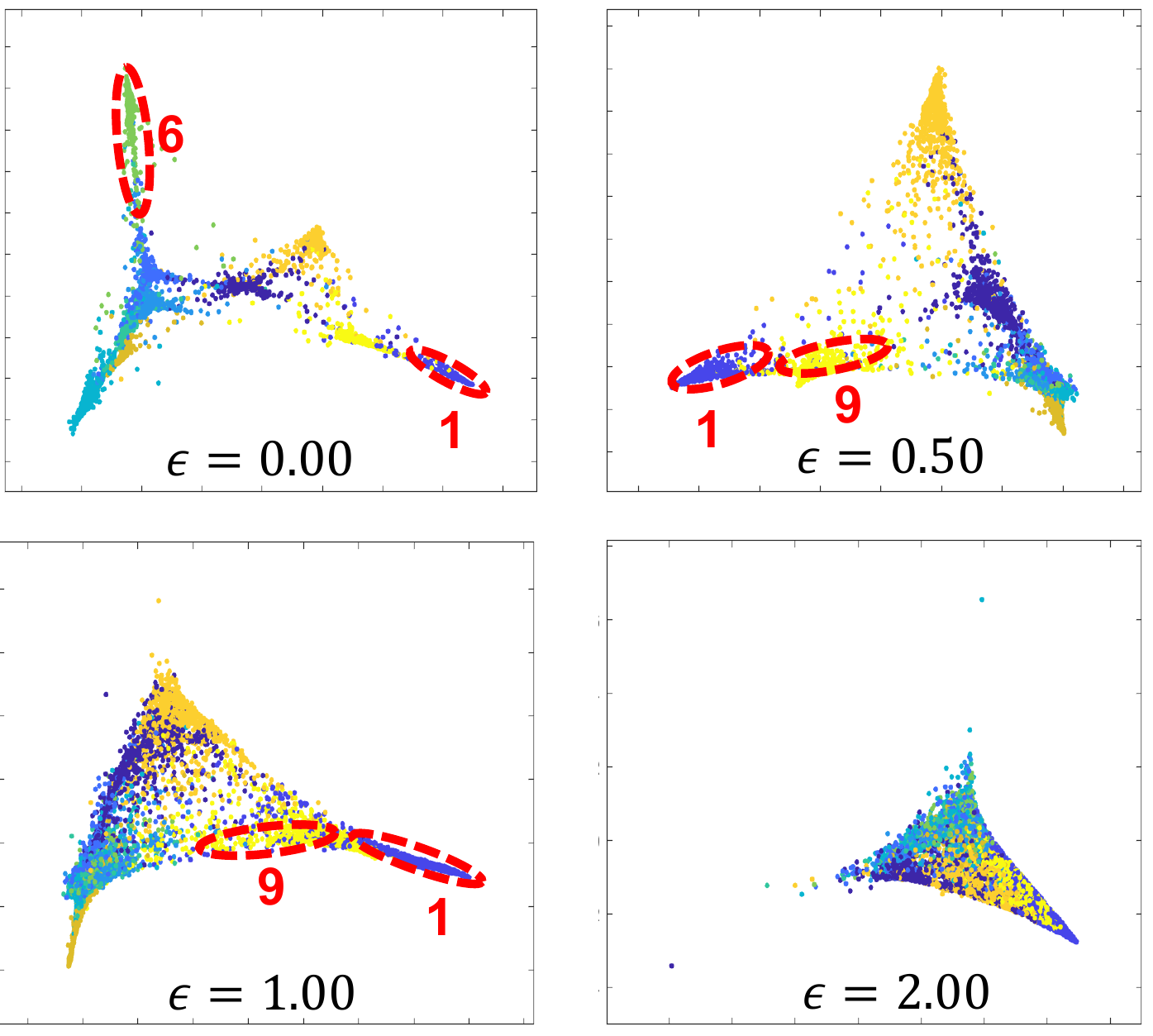}}
\caption{The label SPADE scores and spectral   embeddings  with top two dominant generalized eigenvectors  (CIFAR-10 test set).}
\label{fig:spadeCIFAR}
\end{center}
\vskip -0.2in
\end{figure*}


\begin{figure*}[ht]
\vskip 0.2in
\begin{center}
\centerline{\includegraphics[width=2\columnwidth]{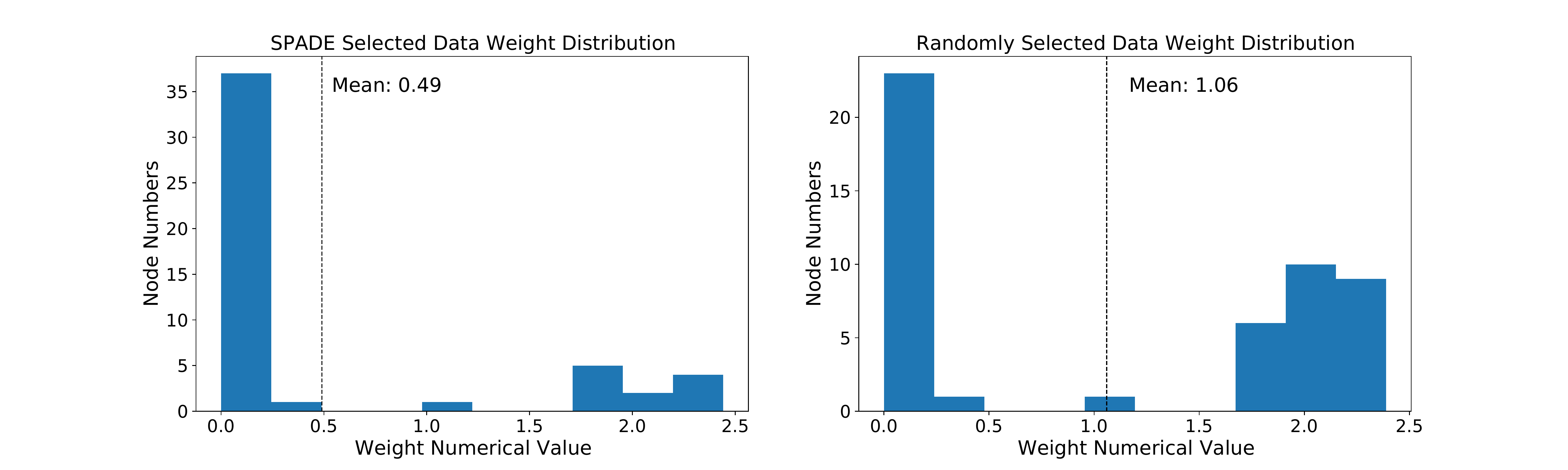}}
\caption{The GAIRAT weight distributions of 50 SPADE-guided (left) and randomly selected (right) data samples}
\label{fig:Geometry}
\end{center}
\vskip -0.2in
\end{figure*}
Figures \ref{fig:spadeMNIST} and \ref{fig:spadeCIFAR} show the SPADE scores of each label \footnote{The SPADE score of each label (class) equals the mean SPADE score of the data samples with the same label.} under four different adversarial robustness levels for the MNIST and CIFAR-10 test sets, where the corresponding 2D spectral embeddings using dominant generalized eigenvectors are also demonstrated. It is observed that for the clean models ($\epsilon=0$), labels $4$  and $9$ ($1$  and $6$) are the most vulnerable labels for MNIST (CIFAR-10).  After adversarial training  with different adversarial robustness levels, labels $1$ and $7$ remain  the most vulnerable ones for MNIST, while labels $1$ (automobile) and $9$ (truck) remain the most vulnerable ones for CIFAR-10.  It is also observed in both test cases that the spectral embeddings become less scattered for more adversarial robust models. As shown in Figure \ref{fig:spadeCIFAR}, the spectral embedding of   the   most robust model trained with $\epsilon=2$   only forms  a single cluster, implying indistinguishable vulnerabilities across different labels.

\subsection{SPADE for Topology Analysis of Neural Networks }
\begin{figure}[htp]
    \centering
    \includegraphics[width=0.991\columnwidth]{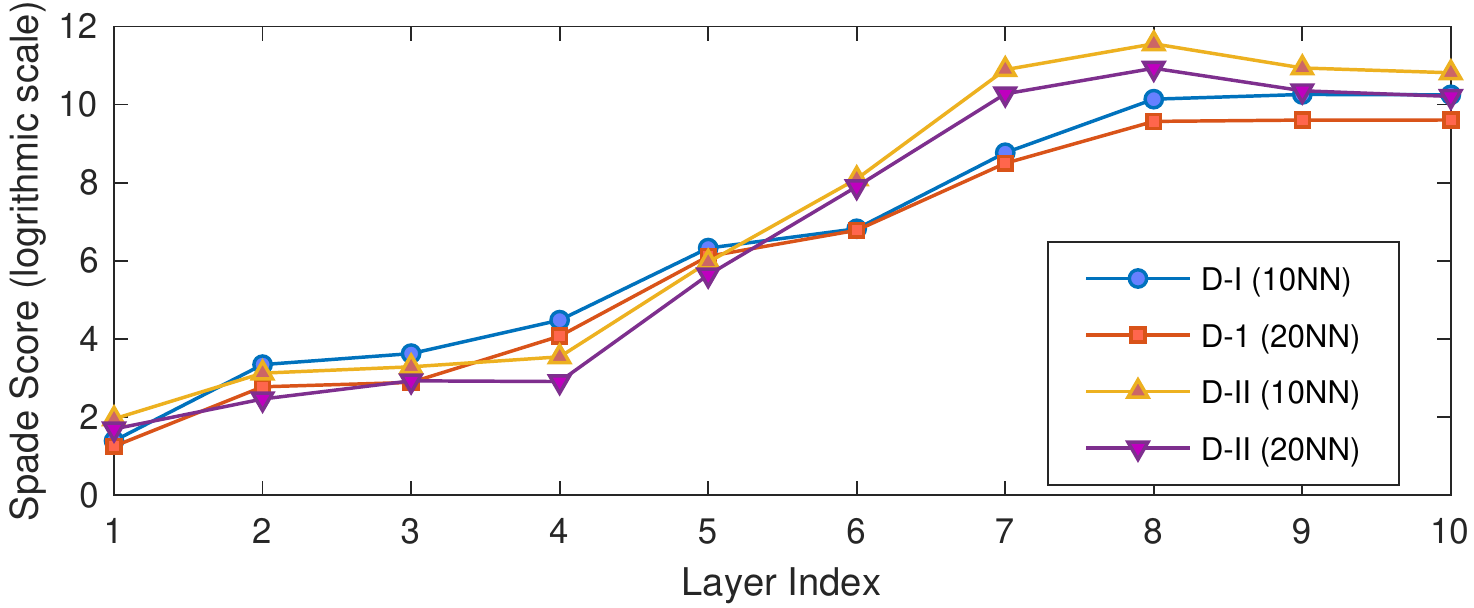}
    \vskip -5pt
    \caption{The layer-wise SPADE scores of  neural network models }
    \label{fig:topnn}

\end{figure}

Figure \ref{fig:topnn} shows the SPADE scores computed using the output data associated with different layers of a neural network model trained on the point cloud datasets D-I and D-II \cite{naitzat2020topology}, which  implies that deeper architectures will result in greater Lipschitz constant and thus SPADE scores. Such results also allows  analyzing the topology changes across  different neural network layers.  For instance, it is  observed that the sharply decreasing Betti numbers for the neural network layers $4$ to $7$ of the  D-II data set   correspond to dramatically increasing SPADE scores.



\subsection{Model SPADE Score for the CINIC-$10$ dataset }
\begin{table*}[ht!]
\caption{Model SPADE scores for the CINIC-10 dataset \cite{darlow2018cinic}  }
\label{cinic_spade}
\begin{center}
\begin{small}
\small	
\begin{sc}
\begin{tabular}{lccccc}
\toprule
CINIC & SPADE(10NN) &  SPADE(20NN) & SPADE(50NN) & SPADE(100NN)\\
\midrule
$\epsilon=0$     & $648.64$ &$ 1393.71$ & $263.69$ &$214.01$\\
$\epsilon=0.25$     & $348.45$ &$ 267.77$ & $201.33$ &$164.30$\\
$\epsilon=0.5$      & $285.68$ &$219.91$ & $166.53$ &$135.66$\\
$\epsilon=1.0$     & $274.17$ &$210.18$ & $156.00$ &$124.55$\\

\bottomrule
\end{tabular}
\end{sc}
\end{small}
\end{center}
\vspace{-10pt}
\end{table*}
Apart from MNIST and CIFAR-$10$ data sets, we further compute the SPADE scores on the CINIC-$10$ dataset~\cite{darlow2018cinic}, which is larger version of CIFAR-$10$ and its test set contains $90,000$ images (data points). The result is demonstrated in Table \ref{cinic_spade}. Similar to the results of MNIST and CIFAR-$10$ data sets, the more robust neural networks always reveal lower SPADE scores. 

\subsection{Model SPADE Score for Natural Robustness}
\begin{table*}[ht!]
\caption{Model SPADE scores for four networks trained with standard CIFAR-10 data and AugMix processed CIFAR-10 data. We evaluate SPADE scores on both clean   test set and corrupted  test set via data shift.}
\label{augmix}
\begin{center}
\begin{small}
\small	
\begin{sc}
\begin{tabular}{lccccc}
\toprule
Networks & AugMix (Clean) &  Standard (Clean) & AugMix (Corrupted) & Standard (Corrupted)\\
\midrule
Allconv     & $149.63$ &$ 213.03$ & $60.68$ &$39.99$\\
Densenet     & $129.49$ &$ 152.07$ & $53.32$ &$33.63$\\
Resnext      & $571.15$ &$614.63$ & $293.94$ &$51.57$\\
WRN     & $738.35$ &$784.03$ & $100.17$ &$72.96$\\

\bottomrule
\end{tabular}
\end{sc}
\end{small}
\end{center}
\vspace{-10pt}
\end{table*}
\cite{hendrycks2019augmix} proposes a data processing technique called AugMix, which enhances the natural robustness of various CNN architectures to data shift. In this experiment, we compute SPADE scores for both standard models and AugMix trained models. As demonstrated in Table\ref{augmix}, the more robust (AugMix-trained) models always have lower SPADE scores on the clean test set, but greater SPADE scores for the corrupted data set generated via adding various noises~\cite{hendrycks2019benchmarking}. 


\subsection{Comparison of Graph Construction Methods}
\begin{table*}[ht!]
\caption{Comparison of  model CLEVER scores on MNIST/CIFAR-10 test sets  \cite{weng2018evaluating}. CNN, DD, and 2CL represent the 7-layer AlexNet-like, Defensive Distillation, and 2-convolutional-layer CNNs. GRASS \cite{feng2020grass}, PCA, CKNN, and KNN stand for different graph-based manifold constructions. ``T10'' (``B10'') denote the SPADE-guided CLEVER scores computed via sampling the top (bottom) $10$ most non-robust samples, respectively, based on node SPADE score.} 
\label{clever_different_G}
\begin{center}
\begin{small}
\small	
\begin{sc}
\begin{tabular}{lccccc}
\toprule
Networks & GRASS (10NN, T10) &  GRASS (10NN, B10)& PCA (10NN, T10) &  PCA (10NN, B10)\\
\midrule

MNIST-2CL($\epsilon = 0$)     & $0.964/0.0465$  &$ 21.136/0.985$ & $ 0.195/0.011$ &$ 21.742/1.078$ \\

MNIST-2CL($\epsilon = 0.3$)   & $0.045/0.006$ &$ 5.574/0.623$ & $ 0.034/0.005$ &$ 7.750/0.971$ \\

MNIST-MLP    & $0.432/0.020$ &$1.729/0.089$ & $0.460/0.022$ &$1.564/0.079$ \\

MNIST-CNN    & $ 0.570/0.044$ &$0.983/0.083$ & $0.525/0.040$ &$1.099/0.099$ \\

MNIST-DD     & $0.448/0.028$ &$1.306/0.094$ & $0.377/0.027$ &$1.487/0.101$ \\

CIFAR-MLP    & $0.113/0.002$ &$0.253/0.005$ & $0.535/0.012$ &$0.214/0.004$ \\

CIFAR-CNN    & $0.230/0.007$ &$0.167/0.004$ & $0.214/0.006$ &$0.114/0.003$\\

CIFAR-DD     & $0.287/0.009$ &$0.111/0.003$ & $0.245/0.008$ &$0.167/0.005$\\

\toprule
Networks &  CKNN (10NN, T10) &  CKNN (10NN, B10)& KNN (10NN, T10)& KNN (10NN, B10)\\
\midrule
MNIST-2CL($\epsilon = 0$)     &  $0.784/0.039$  &$0.801/0.040$ & $0.049/0.002$ &$20.499/0.950$\\

MNIST-2CL($\epsilon = 0.3$)   & $0.541/0.049$  &$0.712/0.072$ & $0.112/0.008$ &$4.888/0.620$\\

MNIST-MLP     & $0.555/0.039$  &$0.850/0.058$ & $1.317/0.067$ &$1.715/0.089$\\

MNIST-CNN    &  $0.070/0.001$  &$0.204/0.004$ & $0.379/0.030$ &$0.894/0.079$\\

MNIST-DD     & $0.045/0.001$  &$0.100/0.002$ & $0.408/0.026$ &$1.399/0.097$\\

CIFAR-MLP    & $0.078/0.002$  &$0.129/0.004$ & $0.213/0.004$ &$0.253/0.005$\\

CIFAR-CNN    & $0.185/0.009$  &$8.938/0.423$ & $0.141/0.004$ &$0.215/0.006$\\

CIFAR-DD     & $0.434/0.048$  &$0.075/ 0.003$ & $0.318/0.036$ &$0.111/0.003$\\

\bottomrule
\end{tabular}
\end{sc}
\end{small}
\end{center}
\vspace{-10pt}
\end{table*}
This set of experiments examine how graph construction methods would impact the computation of SPADE scores. To this end, we have tested the following four methods for constructing the  graph-based manifolds:
1. GRASS: The vanilla kNN graph with spectral sparsification~\cite{feng2020grass};
2. PCA: The vanilla kNN graph constructed after mapping the original data samples to lower dimensional space via principal component analysis (PCA)~\cite{jolliffe2016principal};
3. CKNN: The kNN graph constructed using continuous k-nearest neighbors (CKNN)~\cite{berry2016consistent};
4. KNN: The vanilla k-nearest neighbor graph. 

We apply the above graph construction methods and compute the sample SPADE scores accordingly. Then   $10$ most non-robust (top) and robust (bottom) samples are used for computing the CLEVER scores. Since the CLEVER score is the lower bound on the minimal distortion to obtaining adversarial samples~\cite{weng2018evaluating}, we expect that the CLEVER score computed by robust samples should be greater than that computed by non-robust samples. As shown in Table \ref{clever_different_G}, all four graph construction methods obtain a larger CLEVER score for B10 than T10 for most models. However, the gap between CLEVER scores of T10 and B10 varies from different graphs and models. For instance, the CKNN-based CLEVER score has a relatively smaller gap between T10 and B10 than other methods for the MNIST-2CL model, while it has a larger gap than other methods for the CIFAR-CNN model.


\subsection{Correlation between SPADE and GAIRAT}

\cite{zhang2020geometry} introduces a white-box measurement of data robustness called GAIRAT, which adopts the idea that a data sample closer to the decision boundary is less robust. Consequently, GAIRAT assigns smaller weights on training losses to data samples that are farther from the decision boundary. To verify that our node SPADE score indeed captures the robustness of data samples,  we compare the weights of the top $50$ most robust samples (with the lowest SPADE scores) with the weights of  $50$  randomly selected samples.  As observed in Figure \ref{fig:Geometry}, the SPADE-guided samples have  much smaller weights than the randomly selected samples, which implies the node SPADE scores can be leveraged for effectively identifying the most robust samples. 



\end{document}